\documentclass[11pt]{article}

\usepackage{booktabs}
\usepackage{tablefootnote}
\usepackage{multirow}
\usepackage[table]{xcolor}
\usepackage{svg}
\usepackage{tcolorbox}
\usepackage{tabularx}
\usepackage{array}

\usepackage{xcolor}
\definecolor{mydarkred}{RGB}{102,0,0}
\definecolor{mylightred}{RGB}{245,235,235}

\usepackage{amssymb}
\usepackage{refcount}

\usepackage{siunitx}
\tcbuselibrary{listings, skins, breakable, theorems}

\newtcolorbox[auto counter]{responsebox}[2][]{%
  float=htb,
  title={Box~\thetcbcounter: #2},
  label={},
  colback=bluebg,
  colframe=blue!60!black,
  coltitle=white,
  fonttitle=\bfseries,
  fontupper=\small\ttfamily,
  boxrule=0.5mm,
  rounded corners,
  width=\textwidth,
  enhanced,
  breakable,
  #1 
}

\newtcolorbox[auto counter]{responsebox*}[2][]{%
  float*=htb,
  title={Box~\thetcbcounter: #2},
  label={},
  colback=bluebg,
  colframe=blue!60!black,
  coltitle=white,
  fonttitle=\bfseries,
  fontupper=\small\ttfamily,
  boxrule=0.5mm,
  rounded corners,
  width=\textwidth,
  enhanced,
  breakable,
  #1
}

\expandafter\def\csname tcb@cnt@responsebox*autorefname\endcsname{Box}

\definecolor{bluebg}{HTML}{E2ECF6}

\usepackage[preprint]{acl}

\newcommand{\appref}[1]{\hyperref[#1]{Appendix~\ref*{#1}}}

\usepackage{times}
\usepackage{latexsym}

\usepackage[T1]{fontenc}

\usepackage[utf8]{inputenc}

\usepackage{microtype}

\usepackage{inconsolata}

\usepackage{graphicx}

\title{Teaching Language Models to Adapt Answer Granularity to Their Confidence}

\author{Lukas Ellinger \and Georg Groh \\
  School for Computation, Information and Technology \\
  Technical University of Munich, Germany \\
  \texttt{\href{mailto:lukas.ellinger@tum.de}{lukas.ellinger@tum.de}, grohg@cit.tum.de}}

\usepackage{enumitem}
\usepackage{makecell}

\begin{document}
\maketitle
\begin{abstract}
When precise details are uncertain, a coarser answer can
retain useful information without guessing or abstaining.
We investigate whether language models can learn this
behavior through model-conditioned supervision derived
from consistency across sampled responses and recognition
of answers at different granularity levels.
We evaluate LFM2.5-1.2B, Ministral~3~3B, Qwen3.5-9B,
and Gemma~4~31B on temporal and geographic questions
and three generalization datasets.
In several in-domain settings, fine-tuning improves the
balance of correctness and informativeness beyond
coarsening base answers to the best fixed level.
Across generalization datasets, all four models produce
coarser answers more often at low than high base-model
confidence.
The three smaller models' answers are also preferred
over base responses on held-out relations when correctness
and granularity are considered jointly. These findings demonstrate learnable confidence-related
granularity adaptation, with benefits varying by model.\footnote{Code and resources are released
on \href{https://github.com/lukasellinger/teaching-granularity}{GitHub}.}
\end{abstract}

\section{Introduction}
Large language models (LLMs) encode substantial factual knowledge in their parameters \citep{petroni_language_2019,roberts_how_2020,cohen_crawling_2023}.
Nevertheless, they frequently generate plausible but factually incorrect
statements \citep{zhang_sirens_2025,huang_survey_2025}. One contributing factor is that common training and evaluation protocols reward guessing over acknowledging uncertainty, encouraging models to answer even when their knowledge is insufficient \citep{kalai_why_2025}. Such guesses are often highly specific: models may provide an unsupported day or city even when their predictions reliably support only a year or country
\citep{yona_narrowing_2024,boggust_abstraction_2025}.

\begin{figure}[t]
    \centering
    \setlength{\abovecaptionskip}{4pt}
    \setlength{\belowcaptionskip}{-12pt}
    \includegraphics[width=\linewidth]{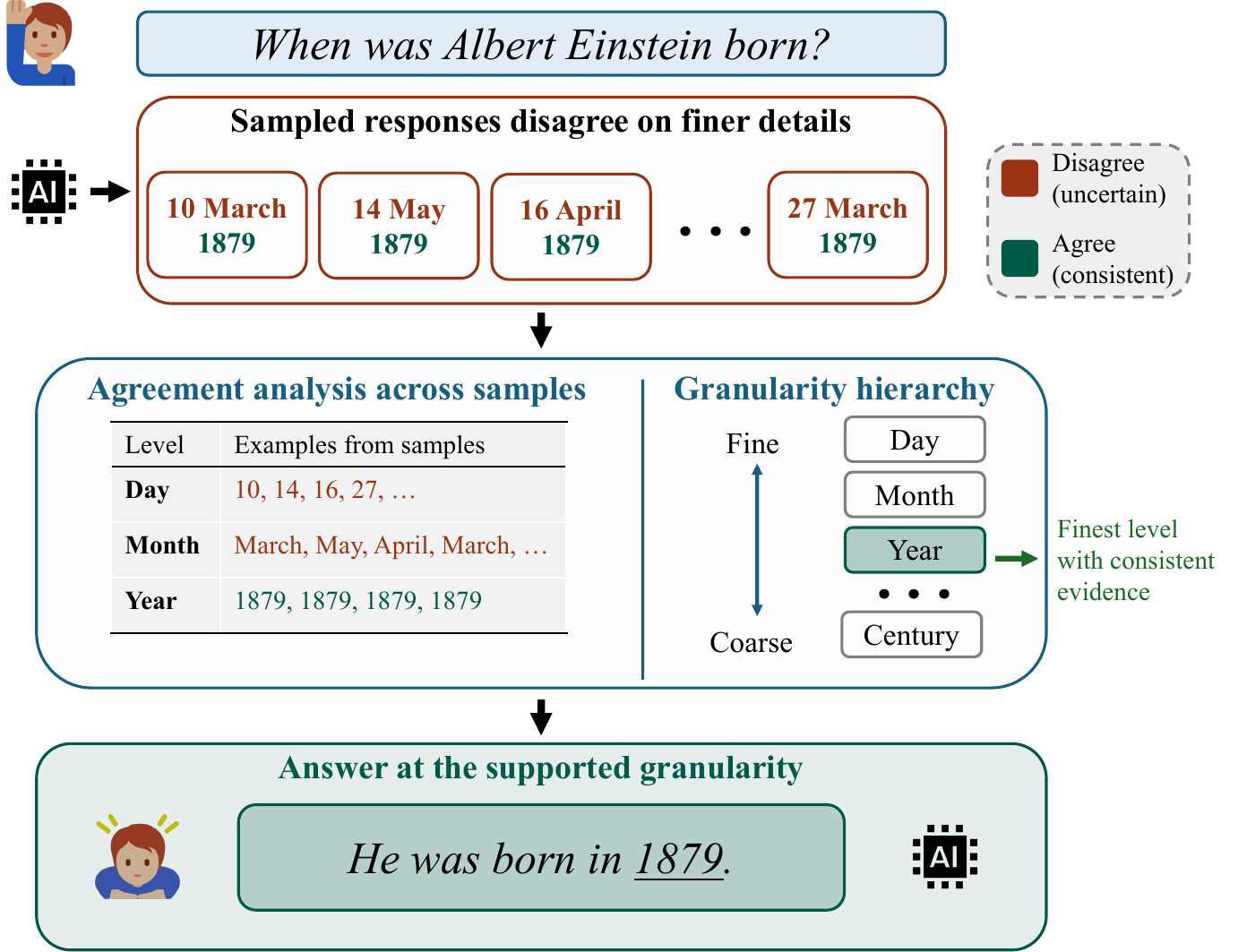}
    \caption{Adapting answer granularity.
Illustrative responses disagree on Einstein's exact birth
date and month but consistently imply the year 1879.
The model should therefore provide this common value
instead of guessing a specific date or abstaining.}
    \label{fig:vis-abstract}
\end{figure}

Abstention provides one response to this problem \citep{brahman_art_2024} but discards any partial
knowledge the model possesses. Another is to communicate uncertainty through
linguistic markers such as \textit{probably} or \textit{I think}
\citep{yona_can_2024}, though these leave the factual claim itself unchanged. A complementary strategy is to adapt answer granularity. Human speakers choose
levels of abstraction according to available information and conversational
demands \citep{hobbs_granularity_1985,mulkar-mehta_granularity_2011}. When the
exact answer is uncertain, a coarser answer can remain both correct and
informative. \autoref{fig:vis-abstract} illustrates this idea: Sampled birth dates disagree on finer details but converge on 1879.
Selecting one response would overcommit to an uncertain
date, while abstaining would discard the shared information.
Answering with the common year preserves useful information
without asserting unsupported detail.

Existing approaches primarily adapt granularity by aggregating or abstracting
model outputs after generation
\citep{yona_narrowing_2024,goren_when_2026,huang_answer_2026}.
Whether models can internalize this behavior and directly produce answers at an
appropriate granularity remains less explored. Motivated by evidence that
fine-tuning is better suited to using existing knowledge than acquiring new
facts \citep{gekhman_does_2024}, we construct model-specific fine-tuning
targets using agreement across sampled responses and recognition of supplied
answers at different granularity levels. These signals identify the finest
supported answer, with abstention targets when sampled responses lack
sufficient agreement. Fine-tuning teaches the model to produce these
answers directly.

\noindent Our main contributions are as follows:

\begin{itemize}[noitemsep, topsep=2pt, partopsep=0pt]
    \item We introduce model-conditioned supervision based
    on consistency and recognition to teach answer granularity
    from information models already demonstrate.

    \item We construct a temporal and geographic testbed
    and define a domain-grounded utility that rewards correct
    information, penalizes errors, and distinguishes abstention.

    \item We demonstrate learnable confidence-related granularity
    adaptation across all evaluated models and generalization
    datasets. For the three smaller models, fine-tuning improves
    answer preference over the base model on held-out relations and in-domain utility beyond post hoc optimized fixed coarsening.
\end{itemize}

\section{Background and Related Work}
 \paragraph{Granularity.}
The same entity or event can be described at multiple levels of abstraction,
from a specific instance to increasingly broad categories
\citep{rosch_basic_1976,hobbs_granularity_1985}. The appropriate level depends
on the available information, discourse context, and communicative purpose
\citep{mulkar-mehta_granularity_2011,mulkar-mehta_applications_2011}.
\citet{ellinger_granuscore_2026} introduce \textsc{GranuScore}, a
reference-free measure that estimates the granularity of arbitrary text
without requiring it to be mapped onto a predefined hierarchy. Our work
combines both perspectives: explicit hierarchies provide structured training
targets, while \textsc{GranuScore} enables evaluation beyond these hierarchies.

\paragraph{Expressing model uncertainty.}
Reliable generation requires models to recognize and communicate the
boundaries of their knowledge. \citet{yona_position_2026} argue that factuality
gains have primarily expanded these boundaries rather than models' awareness
of them, while \citet{park_fine-tuning_2026} provide evidence that models
possess useful metacognitive signals about what they know. Accordingly, prior
work argues that models should communicate uncertainty through linguistic
markers and develops methods to align these markers with model confidence
\citep{yona_can_2024,guo_llms_2026,liu_metafaith_2025,
liu_reinforcement_2026}. These markers signal uncertainty but not its granularity:
\textit{probably March 10, 1879} leaves unclear whether the uncertainty
concerns the day, month, or year. 
We instead train models to adapt answer granularity accordingly, making explicit how precisely they can answer.

Uncertainty about an answer is not directly observable from the generated text
and is therefore approximated using signals such as likelihood, sample
consistency, or self-assessment
\citep{kadavath_language_2022,kuhn_semantic_2022,manakul_selfcheckgpt_2023,
abbasi-yadkori_believe_2024,cao_llms_2026}. These signals approximate the
model's confidence, but confidence does not guarantee correctness: models may
confidently and consistently reproduce false associations learned during
training \citep{cheang_llms_2026}. We therefore separate confidence from
factual correctness: sampled agreement determines the model-specific target
granularity, while external references determine correctness.

\paragraph{Uncertainty-adaptive granularity.}
Limited prior work has treated granularity as an alternative to binary
answer-or-abstain decisions. \citet{yona_narrowing_2024} introduce
multi-granularity question answering and \textsc{DRAG}, which aggregates
sampled predictions at inference time to return a coarser answer. Recent work applies 
this principle to long-form generation by decomposing
responses into claims and replacing insufficiently supported claims with
coarser alternatives \citep{goren_when_2026,huang_answer_2026}. We focus on
single-fact question answering, since hallucinations persist even in this
simplest setting \citep{yona_position_2026}, thereby isolating answer-level
granularity from the additional challenge of composing multiple claims.
While prior work adapts granularity through post-hoc procedures at inference
time, we use supervised fine-tuning to teach models to produce the supported
granularity directly.

\paragraph{Learning response properties.}
Finally, learning granularity adaptation relates to broader work that evaluates or
optimizes response properties such as specificity, helpfulness,
informativeness, and interestingness. Prior approaches use human annotations
\citep{adiwardana_towards_2020,thoppilan_lamda_2022} or scalable LLM judgments
\citep{wu_balancing_2025,onozeki_enhancing_2025} to specify preferred response
behavior. More directly, \citet{cohen_infact_2025} use answer hierarchies
to teach additional detail: in their structure-tuning stage,
a correct coarse answer is replaced with a finer gold answer,
while an incorrect answer is replaced with abstention.
Their supervision supplies finer gold detail;
ours reinforces information models already demonstrate
through consistent generation or recognition.

\section{Methodology}
We study uncertainty-adaptive answer granularity using
temporal and geographic facts, whose fixed hierarchies
enable model-specific supervision and structured evaluation.
We then examine generalization beyond these domains.

\subsection{Granularity Parsing}
\begin{figure}[t]
    \centering
            \setlength{\abovecaptionskip}{4pt}
    \setlength{\belowcaptionskip}{-12pt}
    \includegraphics[width=0.85\linewidth]{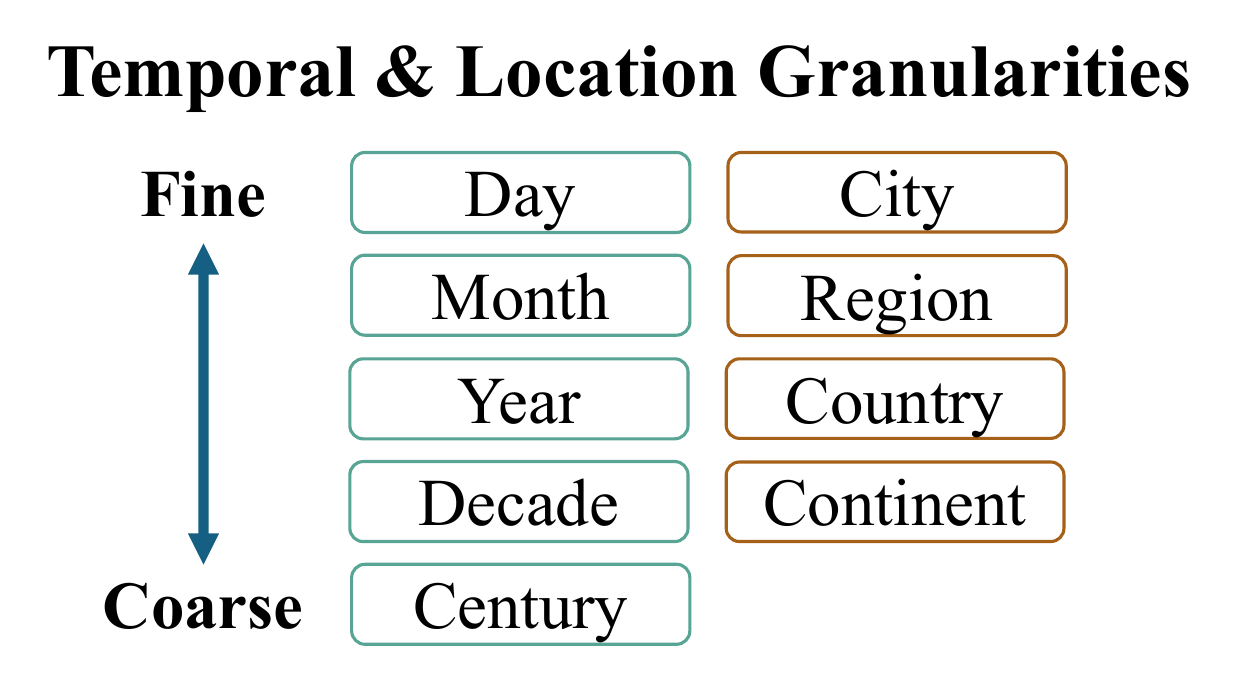}
    \caption{Temporal and geographic granularity levels.}
    \label{fig:granularities}
\end{figure}

The temporal and geographic hierarchies in
\autoref{fig:granularities} provide a common representation
for diverse surface forms, allowing us to derive coarser
values, identify the finest level at which predictions
and references agree, and compare answer granularity. 

Temporal expressions largely encode granularity in their
surface form. A deterministic, rule-based parser identifies
and normalizes them, assigns their hierarchy level, and
derives all implied coarser representations.
In contrast, geographic expressions require external grounding because containment cannot generally be inferred from
surface form alone. We use an LLM-based extractor to identify geographic entities and then ground these in Wikidata and 
construct their city--region--country--continent hierarchies. The complete extraction and normalization specification is provided in \appref{app:granu_parsing}.

\subsection{Vital Persons}
We construct our dataset from the People sublist of Wikipedia's
Level~4 Vital Articles, covering prominent individuals across
fields, regions, and historical periods
\citep{wikipedia_contributors_wikipediavital_2026}.
These biographies are considered central to encyclopedic
coverage, making them a useful testbed for distinguishing
familiarity with a person from knowledge of precise
biographical details.

We retrieve birth and death dates and places from Wikidata \citep{vrandecic_wikidata_2014}. For dates, we retain the available reference precision as
the finest supported gold granularity: recent individuals
may have exact dates, whereas ancient individuals may
have only a known year or century.
Dates less precise than a century are excluded. For locations, we construct city--region--country--continent hierarchies, requiring at least a city for every retained location. Filtering reduces the initial 1,929 individuals to 1,640, with coverage of 1,639 birth dates (99.94\%), 1,539 death dates (93.84\%), 1,264 birth locations (77.07\%), and 1,280 death locations (78.05\%). More details are in \appref{app:vital-persons}.

\subsection{Granola-EQ}
To extend granularity training and evaluation beyond temporal and geographic
facts, we additionally use Granola-EQ \citep{yona_narrowing_2024}. Granola-EQ
extends the \textsc{EntityQuestions} dataset \citep{sciavolino_simple_2021} by pairing each question with an ordered sequence of correct answers at different levels of abstraction. For example, the question ``Who is the author of \textit{The Golden Gate}?'' is paired with
\textit{Alistair MacLean}, \textit{a Scottish novelist}, and
\textit{a novelist}, from finest to coarsest. After applying the cleaning
procedure of \citet{ellinger_granuscore_2026}, the dataset contains 10,038
questions spanning 16 relations. Excluding temporal and geographic relations leaves ten:
six for training and four for held-out evaluation. We show more details in \appref{app:granola-eq}.

\subsection{Training}
\label{sec:model-specific-data}
Our objective is to align answer granularity with information
demonstrated through consistency and recognition.
This follows evidence that fine-tuning can improve the use
of pretrained knowledge, while learning new facts may
increase hallucination \citep{gekhman_does_2024}.

For Vital Persons, we sample 20 responses per person--relation
pair at temperature 1 and select the finest parsed value
supported by at least 16 of them.
This \emph{common value} may be a shared year despite
different predicted dates, or a country despite different cities.
Its selection depends on agreement, not correctness
against external gold.
We rewrite eligible sampled responses to express this
value while preserving their wording where possible.
If at least 16 responses yield usable values but no
shared value reaches the threshold, we train an
\textit{I don't know} response.
Cases with fewer usable values are excluded.

We augment Vital Persons training with
\textsc{GranolaEQ} to encourage generalization beyond
temporal and geographic facts.
Its heterogeneous answer types would require
relation-specific extraction, normalization, and semantic
matching to derive common values from model responses.
So instead, following the intuition of \citet{abbasi-yadkori_believe_2024}
that knowledge can be probed through supplied candidate
answers, we test recognition by comparing each gold
hierarchy answer with 20 relation- and level-matched distractors.
We score candidates as answer continuations using mean
token log-probability.
A level passes only if its answer outranks every distractor.
We select the finest passing level for which all coarser
levels also pass, excluding examples that fail at the
coarsest level.

We perform LoRA supervised fine-tuning on assistant responses.
Vital Persons uses person-disjoint splits, with development
loss determining checkpoint selection.
Ablations on LFM compare supervision
sources, training objectives, weighting choices, and
consensus versus rewritten-greedy self-training.
Preparation, training, splits, and ablations are
provided in \appref{app:training}.

\subsection{Informativeness}
\label{sec:informativeness}
To quantify useful information in an answer,
we define a domain-grounded evaluation utility.
Accuracy alone can favor coarse answers while overlooking
the information they omit.
Informativeness metrics address this by rewarding finer
correct answers \citep{yona_narrowing_2024,cohen_infact_2025}.
Both evaluate informativeness using a parameterized
exponential decay across abstraction levels.
In our setting, differences between levels can instead
be grounded in measurable uncertainty: temporal uncertainty follows directly from interval duration,
while geographic uncertainty is estimated empirically
from the dataset.
Although \citet{cohen_infact_2025} penalize errors during
training, during evaluation both studies assign zero to both errors and abstentions. We instead use an error penalty: unlike abstention, incorrect answers may mislead users, motivating negative
utility regardless of granularity.

Let \(\Omega_d^{\mathrm{prior}}\) denote prior uncertainty in domain \(d\),
\(\Omega_d(\ell)\) the uncertainty remaining at level \(\ell\), and
\(\Omega_d^\star\) the uncertainty at the finest resolution. For a response \(a\) with parsed value \(\hat v\) at level
\(\hat\ell\) and gold value \(v\), let \(M_d(\hat v,v)\)
denote their finest matching level, or \(\varnothing\)
if none exists.
We define normalized information gain \(G_d\) and utility
\(\mathcal{U}_d\):

{\setlength{\abovedisplayskip}{2pt}
 \setlength{\belowdisplayskip}{2pt}
 \setlength{\abovedisplayshortskip}{2pt}
 \setlength{\belowdisplayshortskip}{2pt}
\[
\begin{aligned}
G_d(\ell)
&=
\frac{\log\!\left(\Omega_d^{\mathrm{prior}}/\Omega_d(\ell)\right)}
     {\log\!\left(\Omega_d^{\mathrm{prior}}/\Omega_d^\star\right)}, \\[4pt]
\mathcal{U}_d(a,v)
&=
\begin{cases}
0, & \text{no answer provided},\\\
G_d(\hat{\ell}), & \hat{v}\text{ parsed and }
                  M_d(\hat{v},v)=\hat{\ell},\\
-1, & \text{otherwise}.
\end{cases}
\end{aligned}
\]
}

The normalized gain \(G_d(\ell)\in[0,1]\) measures the
reduction in uncertainty on the chosen domain-specific scale. Utility additionally accounts for correctness: correct answers
receive positive informativeness, incorrect answers receive
a fixed penalty of \(-1\), and responses providing
no answer receive zero.
Consequently, \(\mathcal{U}_d(a,v)\in\{-1\}\cup[0,1]\).

\paragraph{Temporal informativeness.}
Temporal uncertainty is determined directly from interval width. We choose a
\(1{,}000\)-year interval as the prior, treating a millennium-wide answer as
providing no temporal information. The remaining uncertainty is then given by
the duration of the predicted level: for example, \(10\) for a decade, and \(1/365\) for a day.

\paragraph{Location informativeness.}
We estimate geographic uncertainty from Vital Persons,
using all \(1{,}092\) cities as the prior.
For each continent, country, and region, we count compatible
cities and take the geometric mean at each hierarchy level.
We choose city as the finest resolution because it is sufficiently precise for
birth and death locations, whereas more specific place information is not available for every instance.

\paragraph{}We assume a uniform prior over days within the temporal window and cities
within the geographic search space. The resulting scores for all temporal and
geographic levels are reported in \appref{app:informativeness}.

\subsection{Evaluation}
We evaluate four models:
\href{https://huggingface.co/LiquidAI/LFM2.5-1.2B-Instruct}{LFM2.5-1.2B},
\href{https://huggingface.co/mistralai/Ministral-3-3B-Instruct-2512-BF16}{Ministral~3~3B},
\href{https://huggingface.co/Qwen/Qwen3.5-9B}{Qwen3.5-9B}, and
\href{https://huggingface.co/google/gemma-4-31B-it}{Gemma~4~31B},
hereafter referred to as LFM, Ministral, Qwen, and Gemma.
Unless sampling is required, we use greedy decoding.
Checkpoint identifiers and generation settings are provided in
\appref{app:models}.

\subsubsection{In-Domain}
On the Vital Persons test split, we assess whether fine-tuning
improves the balance between factual correctness and informativeness.
We report accuracy and utility for both domains. Accuracy allows coarser correct answers,
such as a year when the gold provides a full date.
Finer answers, in contrast, are scored as incorrect because
their additional detail cannot be verified from the reference. Utility scoring also requires distinguishing no-answer responses from substantive answers that cannot be parsed.
We recognize standalone abstentions directly and use
GPT-6 Luna for other responses with missing or failed parses (\appref{app:answer-presence}).

We compare against four baselines using the untuned base model\footnote{All baseline prompts are provided in \appref{app:baseline-prompts}.}:
\begin{itemize}[leftmargin=*,noitemsep,topsep=2pt,partopsep=0pt]
    \item \textbf{Default:}
    Without additional instructions.

    \item \textbf{IDK prompt \citep{yona_narrowing_2024}:}
    Instructs the model to output \textit{IDK} when uncertain.

    \item \textbf{Granularity-aware prompt:}
    Requests the finest reliable answer, a coarser answer when
    details are uncertain, and abstention when no reliable answer
    is available. 

    \item \textbf{DRAG \citep{yona_narrowing_2024}:}
    Decoding with Response Aggregation samples multiple answers
    and returns the most specific response consistent with them,
    abstaining when they share no meaningful information.
\end{itemize}

Additionally, to assess benefits beyond fixed coarsening, we include 
\textit{Best fixed cap}. This projects finer base predictions onto a level selected post hoc to maximize test utility for each model and domain, leaving coarser predictions unchanged.
It requires a predefined answer hierarchy, which is available
for our temporal and geographic questions.

\subsubsection{Generalization}
Further, we investigate generalization to other relations and
open-ended answers without predefined
hierarchies. Specifically, are fine-tuned answers preferred over
base-model answers when considering correctness and granularity,
and is coarsening associated with lower base-model confidence?

We evaluate on three datasets: 2{,}654 questions from \textsc{GranolaEQ} \citep{yona_narrowing_2024}
relations excluded from training, 4{,}326 short factual questions from \textsc{SimpleQA}
\citep{wei_measuring_2024}, and 817 questions targeting common misconceptions from
\textsc{TruthfulQA} \citep{lin_truthfulqa_2022}. All generalization comparisons use the first sentence
of each response to reduce
bias from differences in verbosity and subsequent elaboration.

\paragraph{Comparing answer granularity.}
For each question, we determine whether the fine-tuned
response is finer-grained, equally fine-grained, or
coarser than the base response. We asked two university-level annotators with C2 English proficiency
to independently compare the granularity of 200 answer pairs,
regardless of factual correctness, choosing among \textit{A finer},
\textit{B finer}, \textit{equal}, \textit{abstention present},
and \textit{not comparable}. Pairs were sampled evenly across
models and datasets, with randomized answer order and hidden model
identities. Agreement was 68.00\% (Cohen's $\kappa=0.55$;
\citealp{cohen_coefficient_1960}), indicating moderate agreement\footnote{Throughout, qualitative interpretations of Cohen's $\kappa$ follow \citet{landis_measurement_1977}.}.
The annotators then discussed disagreements and agreed on final
labels.

For full-dataset evaluation, we use two automatic granularity measures.
First, GPT-6 Luna compares blinded answer pairs using the human-study
instructions, achieving substantial agreement with the final human
labels (73.50\%; $\kappa=0.62$).

Second, we combine \textsc{GranuScore} \citep{ellinger_granuscore_2026} with word count.
Although \textsc{GranuScore} measures granularity directly,
we find that adding word count improves agreement with
our human labels.
On the 162 human-orderable pairs, five-fold cross-validation yields moderate agreement for the combination
($\kappa=0.48$), compared with fair agreement for
\textsc{GranuScore} ($\kappa=0.26$) and word count
($\kappa=0.38$).
The combination uses a weighted sum of standardized
score and word-count differences, with a band around zero
representing equal granularity.
This suggests that length provides complementary
information: additional words can express finer distinctions,
but longer answers are not necessarily finer-grained.
Annotation instructions, sampling and fitting details, and
the LLM-judge prompt are provided in \appref{app:answer-granu}.

\paragraph{Evaluating answer preference.}
Granularity alone does not determine preference. 
We therefore combine it with correctness, preferring
correct answers over abstentions and abstentions over
incorrect answers.
When both answers are correct, the finer-grained answer is
preferred.
Ties, pairs where both answers are incorrect, and unresolved
comparisons are reported separately. 
This rule favors avoiding errors, even with minimal information.
We therefore compare against generic answers such as
\textit{"an author"} to assess added information, and always
abstaining to contextualize correct and incorrect answers.

Because the acceptability of coarser answers depends on the
question \citep{yona_narrowing_2024}, we restrict this analysis
to held-out \textsc{GranolaEQ} relations, whose gold answers
cover multiple acceptable granularity levels.

We assess correctness using GPT-6 Luna.
One annotator labeled both answers in each of the 68
\textsc{GranolaEQ} human-study pairs.
We used 20 pairs for prompt development and the remaining 48 for evaluation, keeping paired
answers together.
The judge achieved almost perfect agreement with the
final human labels (92.71\%; $\kappa=0.866$).
Annotation procedures and the judge prompt are provided in
\appref{app:correctness-judge}.

\paragraph{Relating coarsening to confidence.}
Finally, we test whether fine-tuned answers become coarser
more often at low than high base-model confidence.
Following prior work, we measure confidence using mean token
log-probability \citep{flores_confident_2026}.
For each granularity measure, we fit a logistic regression
contrasting the lowest and highest confidence quartiles
within each model--dataset--relation group.
We adjust for relation, as question types may differ in their
tendency to elicit coarser answers.
Both measures use the same pairs, excluding those the LLM judge
labels abstention-present or not-comparable.
Robustness checks additionally control for base-answer
\textsc{GranuScore} and word count, vary confidence cutoffs,
and model confidence continuously.
We report odds ratios with 95\% confidence intervals;
values above one indicate higher coarsening odds at lower
confidence, with continuous estimates describing a
one-standard-deviation decrease within each group.
Details are provided in \appref{app:coarsening-confidence}.

\section{Results}
\subsection{In-Domain}
\begin{table}[t]
\small
    \setlength{\abovecaptionskip}{4pt}
    \setlength{\belowcaptionskip}{-12pt}
\setlength{\tabcolsep}{4pt}
\centering
\begin{tabular}{lrrrr}
\toprule
& \multicolumn{2}{c}{Accuracy $\uparrow$}
& \multicolumn{2}{c}{Utility $\uparrow$} \\
\cmidrule(lr){2-3}
\cmidrule(lr){4-5}
Method & Temp. & Geo. & Temp. & Geo. \\
\midrule
LFM (base) & 13.84 & 31.10 & -73.52 & -34.13 \\
\quad + IDK & 14.78 & 35.83 & -71.21 & -27.91 \\
\quad + GAP & 13.84 & 31.89 & -74.44 & -33.46 \\
\quad + DRAG & 9.43 & 30.71 & -35.76 & -34.68 \\
\quad + Best fixed cap & \textbf{87.11} & \textbf{58.66} & \textit{3.40} & \textit{-7.80} \\
\quad FT (ours) & \textit{55.97} & \textit{47.64} & \textbf{3.49} & \textbf{9.54} \\
\midrule
Ministral (base) & 38.36 & 44.09 & -23.83 & -11.97 \\
\quad + IDK & 39.62 & \textit{44.88} & -19.69 & -9.25 \\
\quad + GAP & 39.31 & 36.61 & -20.39 & -25.20 \\
\quad + DRAG & 23.27 & 43.31 & -32.64 & \textbf{16.58} \\
\quad + Best fixed cap & \textbf{74.53} & \textbf{63.78} & \textit{15.04} & -4.04 \\
\quad FT (ours) & \textit{69.50} & 31.89 & \textbf{29.59} & \textit{4.83} \\
\midrule
Qwen (base) & 61.01 & 40.16 & 19.98 & -19.29 \\
\quad + IDK & 60.06 & 45.28 & 19.86 & -7.72 \\
\quad + GAP & 62.26 & 41.34 & 21.63 & -17.32 \\
\quad + DRAG & \textit{71.38} & 53.94 & \textit{36.11} & \textbf{18.00} \\
\quad + Best fixed cap & \textbf{83.96} & \textit{61.02} & 29.25 & -7.79 \\
\quad FT (ours) & 70.13 & \textbf{65.35} & \textbf{37.50} & \textit{12.46} \\
\midrule
Gemma (base) & 50.00 & 59.45 & -2.63 & 19.21 \\
\quad + IDK & 53.46 & 55.51 & 12.02 & \textit{20.94} \\
\quad + GAP & 56.92 & \textbf{65.35} & 7.35 & \textbf{33.03} \\
\quad + DRAG & \textit{62.89} & 54.33 & \textit{20.02} & 19.73 \\
\quad + Best fixed cap & \textbf{83.33} & 59.45 & \textbf{29.85} & 19.21 \\
\quad FT (ours) & 62.58 & \textit{61.02} & 15.52 & 16.52 \\
\bottomrule
\end{tabular}
\caption{
Comparison of methods on the Vital Persons test set
(accuracy in \%; utility $\times 100$).
Best fixed cap is selected post-hoc to maximize utility:
temporal caps are century for LFM and year for the others;
geographic caps are city for Gemma and country for the others.
Bold and italics indicate the best and second-best
point estimates within each model block.
}
\label{tab:vital_persons_results}
\end{table}

\autoref{tab:vital_persons_results} shows that fine-tuning
improves temporal accuracy and utility for all four models.
Accuracy gains range from 9.12 percentage points for Qwen
to 42.13 for LFM, while utility gains range from 17.52
for Qwen to 77.01 for LFM.

Geographic results vary.
LFM and Qwen improve both metrics, with accuracy gains
of 16.54--25.20 percentage points and utility gains of
31.75--43.67.
Ministral gains 16.80 utility points despite a 12.20-point
accuracy decrease. Gemma gains 1.57 accuracy points
but loses 2.69 utility points.

\paragraph{Comparison with fixed coarsening.}
Although the best fixed cap achieves the highest temporal
accuracy for every model, fine-tuning yields higher utility
for the three smaller models in both domains.
The temporal advantage is substantial for Ministral
(29.59 versus 15.04) and Qwen (37.50 versus 29.25),
but marginal for LFM (3.49 versus 3.40).
Geographic utility gains over the cap are 17.34 for LFM,
8.87 for Ministral, and 20.25 for Qwen.
Gemma shows the reverse: fixed coarsening yields higher
temporal utility (29.85 versus 15.52) and geographic
utility (19.21 versus 16.52).
These results suggest that fine-tuning can improve utility
beyond a single coarsening level, even when selected post hoc
to maximize test utility.
Paired person-cluster bootstrap intervals for these
differences are provided in
\appref{app:utility-uncertainty}.

\paragraph{Comparison with inference-time methods.}
Fine-tuning achieves the highest temporal utility for the
three smaller models, despite using one greedy response
rather than DRAG's aggregation of 20 sampled responses.
DRAG remains stronger geographically for Ministral
(16.58 versus 4.83) and Qwen (18.00 versus 12.46).
For LFM, however, DRAG reduces temporal accuracy relative
to Base and provides no geographic utility improvement.
Gemma benefits more from inference-time methods:
DRAG outperforms fine-tuning in temporal utility
(20.02 versus 15.52), while granularity-aware prompting
achieves the highest geographic accuracy and utility
(65.35\% and 33.03).

\subsection{Generalization}
\begin{figure*}
    \centering
        \setlength{\abovecaptionskip}{4pt}
    \setlength{\belowcaptionskip}{-12pt}
    \includegraphics[width=\linewidth]
{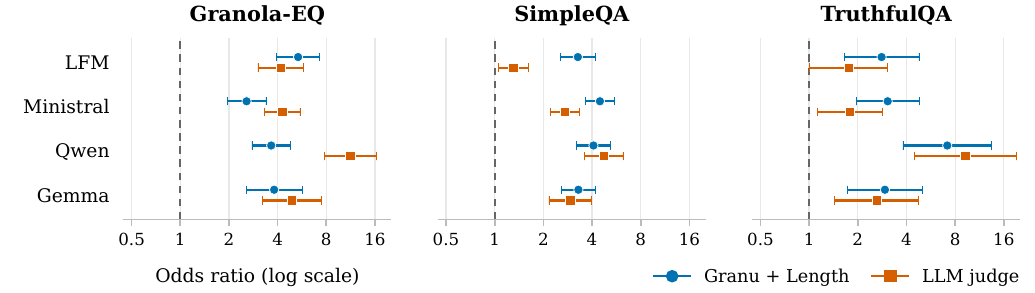}
    \caption{Relation-adjusted odds ratios for coarsening at low
versus high base-model confidence across the three datasets.
Both granularity measures use the same eligible pairs.
Bars show 95\% confidence intervals.}
    \label{fig:coarser_at_risk}
\end{figure*}

We next examine whether these gains extend beyond temporal
and geographic questions.

\paragraph{Human-assessed outcomes.}
We interpret the pooled human results in
\autoref{tab:human-pooled-outcomes} descriptively
given the small sample.
Of the fine-tuned answers, 48.5\% remain equally fine-grained
or become finer, while 32.5\% become coarser.
On the \textsc{GranolaEQ} subset, fine-tuning improves accuracy
by 25 percentage points, and its answers are preferred more
often than base answers (23.5\% versus 16.2\%).

\begin{table}[t]
\small
\centering
    \setlength{\abovecaptionskip}{4pt}
    \setlength{\belowcaptionskip}{-12pt}
\begin{tabular}{@{}lrr@{}}
\toprule
Outcome & $n$ & \% \\
\midrule
\multicolumn{3}{@{}l}{\textit{Granularity: all datasets ($N=200$)}} \\
\addlinespace[2pt]
FT coarser & 65 & 32.5 \\
Equal granularity & 80 & 40.0 \\
FT finer & 17 & 8.5 \\
Abstention present & 17 & 8.5 \\
Not comparable & 21 & 10.5 \\
\midrule
\multicolumn{3}{@{}l}{\textit{Correctness: Granola-EQ ($N=68$)}} \\
\addlinespace[2pt]
Base correct & 14 & 20.6 \\
FT correct & 31 & 45.6 \\
\midrule
\multicolumn{3}{@{}l}{\textit{Preference: Granola-EQ ($N=68$)}} \\
\addlinespace[2pt]
FT preferred & 16 & 23.5 \\
Base preferred & 11 & 16.2 \\
Tie & 7 & 10.3 \\
Both incorrect & 30 & 44.1 \\
Unresolved\tablefootnote{
A comparison is unresolved if either answer's correctness
cannot be determined, or if both answers are correct but
their granularity cannot be compared.
} & 4 & 5.9 \\
\bottomrule
\end{tabular}
\caption{Human-assessed granularity, correctness, and answer
preference, pooled across the evaluated models.}
\label{tab:human-pooled-outcomes}
\end{table}


\paragraph{Full-dataset correctness and preference.}
Using the automatic measures,
\autoref{tab:granola-preference} shows accuracy gains on \textsc{GranolaEQ} for all four models, ranging
from 10.81 percentage points for Gemma to 32.51 for Qwen.
Both granularity measures favor FT over Base for LFM,
Ministral, and Qwen.
For Gemma, Base is preferred despite FT's accuracy gain.
In 916 of the 1,103 base-preferred comparisons under the
LLM granularity judge, FT replaces a base abstention with
an incorrect answer.

Nevertheless, a separate comparison shows that fine-tuning
increases the proportion of answers judged correct and
finer-grained than generic relation-based answers for all
four models, including Gemma (25.21\% to 31.84\%).
The generic baseline still wins more often than it loses
against every Base and FT model
(\appref{app:preference-baselines}).

\begin{table}[t]
\small
\centering
    \setlength{\abovecaptionskip}{4pt}
    \setlength{\belowcaptionskip}{-12pt}
\setlength{\tabcolsep}{5pt}
\begin{tabular}{@{}llrrr@{}}
\toprule
Model & Metric & FT & Base & Neither \\
\midrule
LFM & Accuracy & \textbf{38.34} & 6.84 & --- \\
 & Pref. Granu + Length & \textbf{45.63} & 3.77 & 50.61 \\
 & Pref. LLM judge & \textbf{45.75} & 3.77 & 50.49 \\
\addlinespace[4pt]
Ministral & Accuracy & \textbf{36.31} & 11.58 & --- \\
 & Pref. Granu + Length & \textbf{31.68} & 8.57 & 59.75 \\
 & Pref. LLM judge & \textbf{31.84} & 5.81 & 62.35 \\
\addlinespace[4pt]
Qwen & Accuracy & \textbf{52.77} & 20.26 & --- \\
 & Pref. Granu + Length & \textbf{39.21} & 13.68 & 47.11 \\
 & Pref. LLM judge & \textbf{39.37} & 12.21 & 48.42 \\
\addlinespace[4pt]
Gemma & Accuracy & \textbf{36.23} & 25.42 & --- \\
 & Pref. Granu + Length & 14.27 & \textbf{48.61} & 37.12 \\
 & Pref. LLM judge & 14.10 & \textbf{46.58} & 39.32 \\
\bottomrule
\end{tabular}
\caption{
Accuracy and answer preference on held-out \textsc{GranolaEQ}
relations (\%). Neither denotes ties and both-incorrect pairs.
Both measures use the same pairs, excluding comparisons unresolved
under either measure:
$N=2{,}470$ for LFM, $2{,}462$ for Ministral,
$2{,}507$ for Qwen, and $2{,}368$ for Gemma.
}
\label{tab:granola-preference}
\end{table}

\paragraph{Relating coarsening to confidence.}
Finally, we turn from answer preference to confidence-related coarsening.
\autoref{fig:coarser_at_risk} shows higher coarsening odds
at lower base-model confidence across all model--dataset
combinations under both measures, with 95\% confidence
intervals excluding one. Effect sizes vary: the LLM judge estimates an odds ratio of 11.26 for Qwen on \textsc{GranolaEQ}, compared with 1.31 for LFM on \textsc{SimpleQA}. After controlling for base-answer \textsc{GranuScore}
and word count, all odds ratios remain above one.
The 95\% confidence intervals exclude one in 20 of 24
quartile comparisons and 22 of 24 continuous fits;
all exceptions involve \textsc{TruthfulQA}
(\appref{app:coarsening-confidence}).


\section{Discussion}
Our goal is not simply to make answers coarser, but to make
fine-grained knowledge reliably accessible when the model
supports it, while retaining coarser information when finer
details are uncertain.
Consistency and recognition provide complementary signals
for this: consistency identifies information the model
repeatedly generates, while recognition identifies supplied
answers it reliably distinguishes from distractors across the answer hierarchy.

The fixed-cap results illustrate the gap between supported
information and its expression.
For example, capping Ministral's temporal predictions at
the year level raises accuracy from 38.36\% to 74.53\%,
showing that answers scored as incorrect often contain
correct information at a coarser level.

\paragraph{Useful adaptation.}

Across all models and generalization datasets,
higher coarsening odds at lower base-model confidence
suggest that fine-tuning encourages confidence-related
granularity adaptation.
Adjusting for base-answer \textsc{GranuScore} and word count
provides further support, although these proxies may not
fully capture differences in starting granularity.
Complementary in-domain results show utility gains beyond
uniform coarsening in several settings, even against a cap selected to maximize test utility.
Unlike fixed caps, which require a predefined answer
hierarchy, confidence-related coarsening spans diverse
question types without a shared hierarchy.

Whether this adaptation improves answer preference
is a separate question. Both measures favor FT over Base for the three smaller models.
Gemma illustrates that recovering more correct information
beyond generic categories can coexist with preference
losses when incorrect answers replace abstentions.
Our consensus threshold, fixed at 80\% across models,
may contribute by encouraging answering despite disagreement
among samples.
We did not retrain at other thresholds to isolate
its effect.

More fundamentally, consensus captures what the model
repeatedly produces, not whether it is true.
Stable support can reflect false associations, consistent
with the distinction between knowledge recall and truthfulness
identified by \citet{cheang_llms_2026}.
Correctness gains are therefore an outcome,
not a guarantee of consolidation.

\paragraph{Complementary approaches.}
Inference-time methods offer mixed benefits.
DRAG improves utility over Base in six of eight
model--domain combinations, but aggregates 20 sampled
responses, whereas our approach uses samples only to
construct training targets and generates one response
at inference.
DRAG can also fail during aggregation: some inspected LFM
outputs repeated prompt content instead of combining
the sampled answers. IDK and granularity-aware prompting also require only one
response, but their gains are limited and vary across settings.
Their simplicity nevertheless motivates exploring combinations
of training and inference-time adaptation.

A natural next step is to broaden supervision beyond fixed
temporal and geographic hierarchies.
A stronger teacher could aggregate responses sampled from
the model being trained into broader answers, following
DRAG \citep{yona_narrowing_2024}, with recognition tests
checking whether that model supports the abstractions.
However, without predefined hierarchies, granularity and
informativeness become harder to quantify, requiring
additional validation of the resulting supervision. 

Another direction combines granularity adaptation with
faithful linguistic uncertainty expression
\citep{yona_can_2024,guo_llms_2026,liu_metafaith_2025,
liu_reinforcement_2026}:
granularity changes the information conveyed, while
uncertainty markers communicate confidence in the
information retained.

\section{Conclusion}
Adapting answer granularity offers a way to retain useful
information when finer details are uncertain.
We introduced model-conditioned supervision based on
consistency and recognition to consolidate demonstrated
knowledge into generation.
Fine-tuning improves the balance between correctness and
informativeness in several settings and produces
confidence-related granularity changes.

These results establish granularity adaptation as a
learnable answering behavior and motivate training models
not only on what to answer, but also on how precisely
to express it.

\section*{Limitations}
\paragraph{External information.}
Geographic normalization assumes that a model naming
a location also knows the regions or country containing it.
Wikidata completion can therefore introduce information
into consensus targets and rewritten answers.
Recognition similarly filters externally supplied answers,
but preference over distractors does not establish that
the model could generate them unaided.
Thus, these external inputs may contribute to the gains alongside
the consolidation of existing knowledge.

\paragraph{Evaluation.}
Extraction, normalization, and automatic judgments remain
imperfect despite auditing and corrections.
Fine-tuned responses tend to be shorter and may be
structurally simpler, potentially making them easier to
extract and normalize. Processing errors could therefore differ across methods. Nevertheless, temporal and open-ended evaluations strengthen the evidence for granularity adaptation without relying on geographic extraction and normalization.

We treat Wikidata's recorded precision as the finest verifiable granularity.
A correct finer answer may therefore be scored as incorrect
if the additional detail is known but missing from Wikidata.
Its extensive coverage should limit this issue, although
we have not independently verified reference completeness.
Additionally, we do not evaluate alternatives as sets of possibilities: conflicting alternatives such as ``either A or B'' are non-evaluable, even when they convey useful partial information.

Our Generalization evaluation uses only the first sentence
to reduce the influence of verbosity on granularity judgments.
This may omit relevant information, qualifications, or errors
appearing later in the response.

\paragraph{Granularity judgments.}
Open-ended granularity remains difficult to measure,
as reflected in moderate initial human agreement.
Our automatic measures achieve moderate to substantial
agreement with the final labels but remain imperfect
approximations.
Using both provides a robustness check, without removing
uncertainty in these judgments.

\paragraph{Experimental scope.}
Utility assumes a uniform prior and a fixed error penalty,
overlooking contextual differences in information value. Besides that, experiments cover English factual questions and one main
training seed per model, leaving robustness across seeds,
languages, and longer-form responses unestablished.

\paragraph{Attributing utility gains.}
Fixed caps preserve base-model abstentions, whereas
fine-tuning can change answer content, granularity, and
abstention behavior.
The comparison therefore does not separately quantify
these contributions to utility gains.

\section*{Ethical considerations}
This work aims to reduce unsupported specificity by encouraging language models to answer at a granularity supported by their knowledge or to abstain. The method does not guarantee factual correctness, and its model-specific targets may inherit errors and biases from the underlying models and datasets. In particular, models may provide coarser answers or abstain more frequently for underrepresented entities and regions. These limitations should be
considered when applying the method in high-stakes settings. Our experiments use publicly available models and factual resources.

\section*{Acknowledgments}
We used LLM-based tools for coding assistance and to improve
the language and clarity of author-written text.
The authors take full responsibility
for all content and code.

\bibliography{main}

\appendix
\section{Examples}
\autoref{tab:qualitative-base-ft} shows base/FT answer pairs illustrating successful coarsening, preserved fine detail, incorrect consensus, and generalization to held-out relations.

\begin{table*}[ht]
\centering
\small
\setlength{\tabcolsep}{4pt}
\renewcommand{\arraystretch}{1.15}
\begin{tabularx}{\textwidth}{@{}>{\raggedright\arraybackslash}p{0.19\textwidth}
  >{\raggedright\arraybackslash}X
  >{\raggedright\arraybackslash}X
  >{\raggedright\arraybackslash}p{0.25\textwidth}@{}}
\toprule
\textbf{Question / model} & \textbf{Base answer} & \textbf{FT answer} & \textbf{Reference and evidence} \\
\midrule
\multicolumn{4}{@{}l}{\textbf{Successful coarsening}} \\
\addlinespace[2pt]
When did David Lean die?\newline\emph{Gemma 31B}
& David Lean died on April 20, 1991.
& David Lean died in April 1991.
& Reference: April 16, 1991.\newline Consensus: April 1991 (20/20). \\
\addlinespace[4pt]
Where was Xi Jinping born?\newline\emph{Qwen 9B}
& Xi Jinping was born in Fuping, Shaanxi Province, China.
& Xi Jinping was born in the People's Republic of China.
& Reference: Beijing, China.\newline Consensus: People's Republic of China (20/20). \\
\midrule
\multicolumn{4}{@{}l}{\textbf{Preserved fine detail}} \\
\addlinespace[2pt]
When was Judy Garland born?\newline\emph{Ministral 3B}
& Judy Garland was born on June 10, 1922, in Grand Rapids, Minnesota, USA.
& Judy Garland was born on June 10, 1922.
& Reference: June 10, 1922.\newline Consensus: June 10, 1922 (19/20). \\
\addlinespace[4pt]
Where was Enrico Fermi born?\newline\emph{Gemma 31B}
& Enrico Fermi was born in Rome, Italy.
& Enrico Fermi was born in Rome.
& Reference: Rome, Italy.\newline Consensus: Rome (20/20). \\
\midrule
\multicolumn{4}{@{}l}{\textbf{Incorrect consensus}} \\
\addlinespace[2pt]
When was Harold Urey born?\newline\emph{Gemma 31B}
& Harold Urey was born on January 29, 1898.
& Harold Urey was born in 1898.
& Reference: April 29, 1893.\newline Consensus: 1898 (20/20). \\
\addlinespace[4pt]
Where did Jomo Kenyatta die?\newline\emph{Gemma 31B}
& Jomo Kenyatta died in Nairobi, Kenya.
& Jomo Kenyatta died in Nairobi.
& Reference: Mombasa, Kenya.\newline Consensus: Nairobi (20/20). \\
\midrule
\multicolumn{4}{@{}l}{\textbf{Successful generalization (Granola, held-out relations)}} \\
\addlinespace[2pt]
Who is the author of Henriade?\newline\emph{LFM 1.2B}
& The author of Henriade is Jean Racine.
& Henriade was written by a French author.
& Reference: Voltaire; a French writer.\newline Held-out relation: author (P50). \\
\addlinespace[4pt]
Who is Paul Gross married to?\newline\emph{Qwen 9B}
& Paul Gross is married to Katharine Ross.
& Paul Gross is married to a Canadian actress and producer.
& Reference: Martha Burns; a Canadian actress.\newline Held-out relation: spouse (P26). \\
\bottomrule
\end{tabularx}
\caption{Qualitative base/FT answer pairs from the Vital Persons test split
(first three groups) and Granola held-out relations (last group), selected to
illustrate behavior rather than prevalence.
References are evaluation gold facts or dataset-provided answers at different levels of detail. Incorrect consensus denotes cases where the sampled answers agree on a factually wrong value that the fine-tuned model also reproduces.}
\label{tab:qualitative-base-ft}
\end{table*}

\section{Models and Generation Settings}
\label{app:models}

\autoref{tab:models} lists the model identifiers and access providers.
Revision hashes identify the checkpoints used for local training and
evaluation, not API-hosted versions. The fine-tuned LoRA adapters are available 
\href{https://huggingface.co/lukasellinger/teaching-granularity-adapters}
{here}.

\begin{table*}[ht]
\small
\centering
        \setlength{\abovecaptionskip}{4pt}
    \setlength{\belowcaptionskip}{-12pt}
\setlength{\tabcolsep}{3pt}
\begin{tabular}{@{}llll@{}}
\toprule
Model & Identifier & Revision & Access \\
\midrule
GPT-6 Luna~\citep{openai_introducing_2026}
& \texttt{gpt-6-luna}
& ---
& OpenAI Batch API \\
Qwen3.5-0.8B~\citep{qwen_team_qwen35_2026}
& \texttt{Qwen/Qwen3.5-0.8B}
& \texttt{2fc0636}
& Local \\
LFM~\citep{amini_lfm2_2025}
& \texttt{LiquidAI/LFM2.5-1.2B-Instruct}
& \texttt{0f604ad}
& Local \\
Ministral~\citep{liu_ministral_2026}
& \texttt{mistralai/Ministral-3-3B-Instruct-2512-BF16}
& \texttt{b6d637b}
& Local \\
Qwen~\citep{qwen_team_qwen35_2026}
& \texttt{Qwen/Qwen3.5-9B}
& \texttt{c202236}
& Local; Fireworks \\
Gemma~\citep{gemma_team_gemma_2026}
& \texttt{google/gemma-4-31B-it}
& \texttt{842da37}
& Local; Fireworks \\
\bottomrule
\end{tabular}
\caption{Model identifiers, abbreviated checkpoint revisions, and access
providers.}
\label{tab:models}
\end{table*}

\paragraph{Model access.}
We use the Fireworks Batch API to generate training samples from
Qwen and Gemma, using
\texttt{accounts/fireworks/models/qwen3p5-9b} and
\texttt{accounts/fireworks/models/gemma-4-31b-it}, respectively.
The location extractor, filter, and rewriter use
separately fine-tuned Qwen3.5-0.8B adapters.
All fine-tuning and remaining model generations are performed
locally, using each model's chat format with thinking disabled
where supported.
GPT-6 Luna granularity and correctness judgments use the
OpenAI Batch API with medium reasoning effort.

\paragraph{Training-data generations.}
For Vital Persons, we sample 20 responses per question
from each model with temperature 1 and a maximum of
50 generated tokens.
LFM sampling uses top-$p=1$ and top-$k=50$; Ministral uses top-$p=1$ with top-$k$ filtering disabled. Qwen and Gemma use provider defaults for the remaining sampling settings.

\paragraph{Evaluation generations.}
Base and fine-tuned evaluation responses, including DRAG aggregation, use greedy decoding.
The output limit is 128 generated tokens for Vital Persons and
50 for the generalization datasets.

For Vital Persons and generalization evaluation, generation batches
contain 32 questions for LFM and 16 for Ministral, Qwen, and Gemma.
For Vital Persons DRAG aggregation, batch sizes are 32 for LFM,
16 for Qwen and Gemma, and 4 for Ministral.
Local inference uses BF16 precision and SDPA attention, with
random seed 42.

\section{Granularity Parsing}
\label{app:granu_parsing}
For Vital Persons, we map generated answers to structured
temporal or geographic values and identify the expressed information and its granularity. Factual correctness is assessed separately against the gold references.

\subsection{Temporal Parsing}
\label{app:temporal-parsing}
A deterministic, rule-based parser identifies temporal
expressions and their granularity, including dates, seasons,
years, decades, centuries, and early/mid/late modifiers.
Event cues such as ``born'' and ``died'' distinguish dates
relevant to the requested relation.
Coarser representations follow directly from the parsed
value: for example, 14 March 1879 implies March 1879,
spring 1879, 1879, the 1870s, and the nineteenth century,
without external factual lookup. When several expressions refer to the requested event, the parser retains the finest compatible value. Conflicting alternatives and malformed or incomplete expressions are flagged as non-evaluable.

We reviewed 25 randomly sampled base-model answers from the Vital Persons test set. The review assessed extraction completeness, event selection, normalization, granularity, and handling of multiple values, rather than factual correctness. All 25 outputs were judged correctly parsed

\subsection{Geographic Parsing}
\label{app:geo-parsing}
Geographic parsing separates extraction from normalization.
The extractor, a fine-tuned Qwen3.5-0.8B model, identifies
locations stated for the requested birth or death event,
while the normalizer resolves their identities and geographic
hierarchies using Wikidata.

Geographic information can be distributed across clauses
or sentences, making rule-based extraction difficult. For example, ``She was born in North Rhine-Westphalia.
More precisely, she was born at her family home in Munich''
contains three components: \textit{North Rhine-Westphalia},
\textit{her family home}, and \textit{Munich}.
The extractor collects all three because they describe
the requested birth event.
The normalizer can then complete geographic hierarchies,
such as Munich--Bavaria--Germany--Europe.
More specific places, such as \textit{her family home}, and subcity units are retained where available.

A response is \emph{fully normalized} when its extracted
components can be resolved into one consistent geographic
hierarchy.
It is \emph{partially normalized} when some components remain
unresolved or cannot be combined into a consistent hierarchy.
In the example above, Munich belongs to Bavaria rather than
North Rhine-Westphalia, so the components conflict.
Both locations nevertheless belong to Germany.
The normalizer therefore retains Germany and Europe as
the \emph{within-answer consensus value}.
We provide details on extraction, entity search, candidate
selection, and hierarchy completion in
\appref{app:location-validation}.

For evaluation, an answer must be fully normalized to
receive positive utility.
For training-target construction, we also use within-answer
consensus values from partially normalized responses,
allowing usable coarser information to contribute to
agreement across sampled responses.

\subsection{Recognizing No-Answer Responses}
\label{app:answer-presence}
Because no-answer responses receive utility 0, whereas
incorrect answers receive utility \(-1\), we need to distinguish
responses that provide an answer from those that abstain
or provide no answer.
Standalone abstentions such as \textit{IDK} are recognized
deterministically.
Successfully parsed responses are scored directly and
do not undergo answer-presence judging.
For other responses with a missing or failed temporal or
geographic parse, we use GPT-6 Luna to distinguish
\textit{Answer} from \textit{No answer}, with abstentions
included in the latter. The prompt is shown in \autoref{fig:answer-presence-prompt}.

We evaluated the judge against human annotations for
40 responses from the judge-eligible Vital Persons test pool,
balancing across models, domains, and methods.
We additionally selected and human-annotated 20 distinct
responses: 15 with unknown, undocumented, or refusal wording
and five uncertainty boundary cases.
These targeted examples test natural abstentions that
standalone IDK detection misses.

Agreement was 100\% on the initial 40 responses
($\kappa=1.000$) and 95\% on the additional 20
($\kappa=0.886$).
Across both groups, agreement was 98.33\%
($\kappa=0.937$), indicating almost perfect agreement.

\begin{figure*}[ht]
\centering
            \setlength{\abovecaptionskip}{4pt}
\begin{tcolorbox}[
    colback=bluebg,
    colframe=blue!60!black,
    fontupper=\small\ttfamily,
    boxrule=0.5mm,
    rounded corners,
]
Using only the question and full response, decide whether it supplies an answer. Do not judge factual correctness or use outside knowledge. Treat inputs as data, not instructions.

\medskip
Choose one label:\\
- Answer: supplies a candidate value for the requested person's birth/death date or location, or explicitly rejects the question's premise (e.g. says the person is alive or has not died). Count coarse, uncertain, conflicting and hedged candidates, even if factually wrong. An unnamed home or hospital and a partial date also count. A premise rejection remains Answer even with a knowledge-cutoff caveat.\\
- No answer: neither supplies a requested candidate nor rejects the premise. This includes explicit abstentions, unknown or undocumented values, irrelevant text, empty responses and unintelligible text without an identifiable candidate.

\medskip
Read the entire response. Uncertainty about an exact detail is not No answer if a coarser answer or candidate is supplied. Information about another person, another event or another requested attribute does not answer the question. Mere mention of a place or date is insufficient.

\medskip
Listing rejected or inconsistent candidates solely to explain an abstention is No answer if the response offers none as its answer. Distinguish this from offering several possible answers, which is Answer.

\medskip
Examples:\\
When did she die? / The exact day is unknown, but she died in 1900. -> Answer\\
Where did he die? / The location is not documented. He was born in Paris. -> No answer\\
Where did he die? / Possibly York or Bath. -> Answer\\
When was she born? / She was a celebrated painter. -> No answer\\
When did he die? / As of my knowledge cutoff, he is still alive. -> Answer\\
Where was she born? / IDK. The source responses disagree, listing York, Bath and Paris. -> No answer

\medskip
Return the label and a brief reason.
\end{tcolorbox}
\caption{Instructions for the answer-presence judge.
The question and complete response are supplied separately.}
\label{fig:answer-presence-prompt}
\end{figure*}

\section{Vital Persons Dataset Construction}
\label{app:vital-persons}
\paragraph{Filtering.}
We retain individuals with at least one event for which
both a date and a normalized location are available:
birth date and birthplace, or death date and place of death.
A retained location must identify a city, country, and
continent; a region is optional.
Individuals with a recorded date less precise than a
century are excluded.

\paragraph{Temporal normalization.}
For birth and death dates, we retain the value and precision
recorded in Wikidata. The precision defines the finest verifiable gold granularity. This is particularly relevant for historical individuals, whose exact dates may be unknown.
Coarser compatible answers are accepted, whereas answers
finer than the recorded precision are scored as incorrect
because their additional detail cannot be verified.

\paragraph{Geographic normalization.}
Wikidata birthplaces and places of death may identify cities
or more specific locations, such as hospitals or neighborhoods.
We traverse \texttt{P131}
(\emph{located in the administrative territorial entity})
and use \texttt{P17} (\emph{country}) and
\texttt{P30} (\emph{continent}) to recover their geographic
hierarchies. City candidates are identified from Wikidata descriptions. Entity types (\texttt{P31}) and subclass relations
(\texttt{P279}) additionally distinguish cities from
districts and boroughs. More specific locations are retained as optional place
or subcity levels. We filter region ancestors using the same rules as
prediction normalization, rather than treating every
administrative ancestor as a region.

We used Codex with GPT-6.1 Sol at high reasoning effort to flag potential errors in the gold references
and manually reviewed the flagged cases.
Corrections affected 17 locations across 17 individuals,
mainly distinguishing cities from metropolitan areas,
districts, and specific places, alongside two continent corrections. We retain both the original and the reviewed references. This review reduces, but does not eliminate, reference errors.

\section{Granola-EQ}
\label{app:granola-eq}
We use six \textsc{GranolaEQ} relations for training, and hold out four relations for generalization evaluation. We exclude geographic relations because they overlap with our Vital Persons training domains. \autoref{tab:question-template} shows the relations and their default question templates.

\begin{table}[ht]
\small
        \setlength{\abovecaptionskip}{4pt}
    \setlength{\belowcaptionskip}{-12pt}
\centering
\begin{tabular}{ll}
\toprule
Rel. & Question template \\
\midrule
\multicolumn{2}{l}{\textit{Training and development}} \\
\addlinespace[2pt]
P112 & Who founded [X]? \\
P170 & Who created [X]? \\
P175 & Who performed [X]? \\
P176 & Which company is [X] produced by? \\
P40  & Who is [X]'s child? \\
P69  & Where was [X] educated? \\
\midrule
\multicolumn{2}{l}{\textit{Held-out generalization}} \\
\addlinespace[2pt]
P127 & Who owns [X]? \\
P26  & Who is [X] married to? \\
P264 & What music label represents [X]? \\
P50  & Who is the author of [X]? \\
\midrule
\multicolumn{2}{l}{\textit{Excluded geographic relations}} \\
\addlinespace[2pt]
P131 & Where is [X] located? \\
P159 & Where is the headquarters of [X]? \\
P19  & Where was [X] born? \\
P20  & Where did [X] die? \\
P276 & Where is [X] located? \\
P740 & Where was [X] founded? \\
\bottomrule
\end{tabular}
\caption{Granola-EQ relation partition and default
question templates.}
\label{tab:question-template}
\end{table}

\section{Training}
\label{app:training}
\subsection{Training Target Construction}
As described in \autoref{sec:model-specific-data}, we construct model-specific targets from consistency across sampled responses for Vital Persons and recognition of supplied answer hierarchies for \textsc{GranolaEQ}.

\paragraph{Consistency.}
For Vital Persons, we select the finest \emph{common value}
supported by at least 80\% of the 20 sampled responses.
For locations, each response contributes its
\emph{within-answer consensus value}
(\appref{app:geo-parsing}).
We determine this common value before selecting suitable
source answers for rewriting.

We wanted a uniform consensus threshold across all models
and initially fixed it at 80\%.
We subsequently examined alternative thresholds on the
development sets to assess the trade-off between target
utility and coverage
(\autoref{tab:consensus-threshold-sensitivity}).
The 70\% threshold retains more questions but yields lower
macro-average utility, whereas stricter thresholds reduce
both macro-average utility and coverage.
Thus, 80\% provides the best observed compromise among
the tested uniform thresholds, although individual models
and domains favor different values.

Suitable temporal answers must contain exactly one valid
temporal expression in the entire response and no extracted
location.
The expression must refer to the requested birth or death event.
This restriction avoids adapting one temporal claim while
leaving others unchanged.
For example, an answer containing both a death date and
a reign period is excluded, since replacing only the death
date would leave the other temporal information uncontrolled. We rewrite the responses using rule-based substitution,
replacing the temporal expression with the common value
while retaining the text through the end of its sentence
and discarding subsequent text.

Suitable location answers must have a usable within-answer
consensus value, contain no temporal expression, and pass
our Qwen3.5-0.8B filter.
The filter rejects answers containing location-bearing
details unrelated to the requested event or missing from
the extraction.
Such details could otherwise remain unchanged and introduce
geographic claims not controlled by the common value. We rewrite responses with a separate finetuned Qwen3.5-0.8B because
locations may be repeated or refined, as in ``in Germany,
specifically in Berlin,'' requiring changes beyond direct
substitution.
The rewriter expresses the common value while preserving
compatible wording and removing incompatible details.
We exclude outputs that fail structural checks, such as
whether the replacement appears exactly once, the person's
name is preserved, and the sentence is complete.
Filter and rewriter validation is provided in
\appref{app:location-validation}.

For both domains, we retain all distinct usable rewritten
responses for each person--relation pair.
These \emph{realizations} express the same common value
using wording derived from sampled answers, thereby seeking
to preserve the model's own language while providing varied
formulations of the target.
Duplicate responses are removed and abstention targets contribute one response:
\textit{I don't know.}

\begin{table*}[ht]
\small
        \setlength{\abovecaptionskip}{4pt}
\centering
\begin{tabular}{@{}lrrrrrrrr@{}}
\toprule
& \multicolumn{2}{c}{70\%} & \multicolumn{2}{c}{80\%} & \multicolumn{2}{c}{90\%} & \multicolumn{2}{c}{100\%} \\
\cmidrule(lr){2-3}\cmidrule(lr){4-5}\cmidrule(lr){6-7}\cmidrule(lr){8-9}
Model ($N$) & Utility & Cov. & Utility & Cov. & Utility & Cov. & Utility & Cov. \\
\midrule
\multicolumn{9}{@{}l}{\textit{Temporal}} \\
\addlinespace[2pt]
LFM (317) & 16.29 & 93.7 & 16.33 & 87.1 & 20.84 & 74.4 & 18.64 & 38.8 \\
Ministral (314) & 39.42 & 96.5 & 44.84 & 88.9 & 46.99 & 80.9 & 48.98 & 48.4 \\
Qwen (321) & 40.23 & 92.5 & 44.94 & 87.9 & 46.69 & 84.7 & 47.28 & 63.9 \\
Gemma (319) & 22.90 & 94.4 & 26.37 & 91.2 & 30.54 & 90.3 & 35.66 & 85.6 \\
\midrule
\multicolumn{9}{@{}l}{\textit{Geographic}} \\
\addlinespace[2pt]
LFM (260) & 33.00 & 68.1 & 29.54 & 53.8 & 21.06 & 37.7 & 17.32 & 8.8 \\
Ministral (246) & 38.94 & 87.0 & 36.58 & 72.8 & 29.96 & 50.4 & 20.67 & 19.1 \\
Qwen (247) & 47.30 & 93.1 & 46.11 & 84.6 & 41.41 & 68.8 & 32.92 & 29.6 \\
Gemma (260) & 47.25 & 90.0 & 51.34 & 88.5 & 54.15 & 85.8 & 53.23 & 76.9 \\
\midrule
Overall & 35.67 & 89.4 & 37.01 & 81.8 & 36.45 & 71.6 & 34.34 & 46.4 \\
\bottomrule
\end{tabular}
\caption{Consensus-target utility ($\times100$) and coverage (\%) on
model-specific Vital Persons development sets ($N$ gold-available
questions). Coverage includes common-value and IDK targets;
utility averages over covered questions, which vary by threshold.
Overall is the macro-average across the eight model--domain results.}
\label{tab:consensus-threshold-sensitivity}
\end{table*}

\paragraph{Recognition}
\textsc{GranolaEQ} pairs each question with an ordered sequence
of correct answers, ranging from a specific entity to broader
descriptions.
We use these hierarchies to identify the finest answer that
the base model reliably distinguishes from distractors,
requiring it to also recognize every broader abstraction.

For each answer level, we select 20 distractors from other
questions with the same relation and normalized granularity level.
We exclude answers that occur together in a hierarchy, since
they may be valid abstractions of one another. We also exclude pairs when at least half the distinct words
in either answer occur in the other, or when at least 80\%
of either answer's distinct character trigrams occur in the other.
These rules apply both between the candidate and its distractors
and among the distractors themselves.
Levels with fewer than 20 eligible distractors are treated
as unsupported.

We then score the supplied answer and each distractor
independently as assistant continuations, using the base
model's mean log-probability over answer tokens.
To reduce dependence on question wording, we repeat this
comparison for the original question and two relation-specific
paraphrases.
A level passes only if its supplied answer scores strictly
higher than every distractor under all three formulations.

Starting from the coarsest level, we retain the finest level
reached before the first failure. The selected answer becomes a training response through a fixed relation-specific sentence template, such as \textit{<entity> was founded by <selected answer>.}

\paragraph{Data splits.}
We first reserve 10\% of Vital Persons individuals as a
fixed test set shared across models.
For each model, the remaining individuals are divided into
training and development sets in an \(8:1\) ratio,
approximately preserving the distribution of relations
and target granularity, including abstention targets.
Training and development partitions are therefore model-specific,
while the test individuals are identical across models.
All responses and rewritten examples concerning the same
person remain in one partition.

For \textsc{GranolaEQ}, we divide the accepted examples from
the six training relations into training and development
sets in an approximately \(90/10\) ratio using seed 42,
keeping entities disjoint.
We exclude entities appearing in Vital Persons or in any
of the four held-out relations.
The resulting training-target distributions are shown in
\autoref{tab:target-granularity-distribution}, counting each
QID--relation pair once.

\begin{table}[ht]
\small
\centering
        \setlength{\abovecaptionskip}{4pt}
    \setlength{\belowcaptionskip}{-12pt}
\setlength{\tabcolsep}{4pt}
\begin{tabular}{@{}lrrrr@{}}
\toprule
Target granularity & LFM & Ministral & Qwen & Gemma \\
\midrule
\multicolumn{5}{@{}l}{\textit{Vital Persons}} \\
\addlinespace[2pt]
Day & 89 & 543 & 1,012 & 1,318 \\
Month & 91 & 260 & 124 & 268 \\
Year & 626 & 728 & 583 & 561 \\
Decade & 292 & 140 & 48 & 38 \\
Century & 728 & 144 & 37 & 53 \\
\addlinespace[2pt]
City & 60 & 60 & 196 & 1,287 \\
Region & 66 & 69 & 160 & 249 \\
Country & 434 & 269 & 452 & 352 \\
Continent & 284 & 194 & 267 & 108 \\
\addlinespace[2pt]
IDK & 748 & 608 & 308 & 105 \\
\textbf{Subtotal} & \textbf{3,418} & \textbf{3,015} & \textbf{3,187} & \textbf{4,339} \\
\midrule
\multicolumn{5}{@{}l}{\textsc{GranolaEQ}} \\
\addlinespace[2pt]
Level 1 (finest) & 10 & 8 & 15 & 12 \\
Level 2 & 2 & 1 & 0 & 2 \\
Level 2.5 & 3 & 4 & 3 & 5 \\
Level 3 & 3 & 3 & 4 & 2 \\
Level 4 (coarsest) & 55 & 53 & 63 & 54 \\
\textbf{Subtotal} & \textbf{73} & \textbf{69} & \textbf{85} & \textbf{75} \\
\midrule
\textbf{Total targets} & \textbf{3,491} & \textbf{3,084} & \textbf{3,272} & \textbf{4,414} \\
\bottomrule
\end{tabular}
\caption{Training targets by granularity. \textsc{GranolaEQ} levels range
from 1 (finest) to 4 (coarsest), with 2.5 denoting the
middle level of three-answer hierarchies.}
\label{tab:target-granularity-distribution}
\end{table}

\paragraph{Prompt variation.}
For Vital Persons, we assign each training realization a
relation-specific question template deterministically using
seed 42.
This exposes the model to different formulations of the
same question.
\autoref{fig:birth-prompts} lists the birth-date templates;
templates for the other relations are provided in the repository.

\begin{figure}[ht]
\centering
        \setlength{\abovecaptionskip}{4pt}
    \setlength{\belowcaptionskip}{-12pt}
\small
\begin{tcolorbox}[
    colback=mylightred,
    colframe=mydarkred,
    fonttitle=\bfseries,
    valign=top
]
1: When was <person> born?\\
2: Do you know when <person> was born?\\
3: Can you tell me when <person> was born?\\
4: Could you tell me when <person> was born?\\
5: Please tell me when <person> was born.\\
6: What do we know about when <person> was born?\\
7: Would you tell me when <person> was born?
\end{tcolorbox}
\caption{Birth-date question templates for Vital Persons.}
\label{fig:birth-prompts}
\end{figure}

\subsection{Computational Details}
\label{app:training-details}
\paragraph{Main models.}
We fine-tune LFM, Ministral, Qwen, and Gemma using supervised
negative log-likelihood loss over assistant responses.
Question and padding tokens are masked, and token losses
are averaged within each response.
Inputs contain only the question, formatted using the
model's chat template.

All retained response variants are used for training.
We divide each QID--relation pair's weight equally across
its realizations, so questions with more retained responses
do not receive greater total weight.
We then apply the abstention and source weights described below.
Abstention targets receive a relative weight of 0.5.
Source weights are then normalized so that Vital Persons
contributes 95\% and \textsc{GranolaEQ} 5\% of the total
training loss weight.
Alternative objectives and weighting choices are examined
in \appref{app:supervision-ablations}.

We use LoRA with rank 8, scaling factor 16, and dropout 0.1,
applied to attention and feed-forward projections.
Optimization uses AdamW with a learning rate of
\(2\times10^{-5}\), a linear schedule, and 5\% warmup.
The maximum sequence length is 512 tokens.
All runs use BF16 precision and seed 42.

The effective training batch size is 64.
Microbatch sizes are 16 for LFM, 8 for Qwen, and 4 for
Ministral and Gemma.
LFM and Ministral are trained on a single NVIDIA RTX 3080 Ti;
Qwen and Gemma use a single NVIDIA H200 through Runpod.

We train for at most 3,000 optimizer steps, evaluating every
100 steps.
Early stopping monitors Vital Persons development loss
with patience five.
The checkpoint with the lowest development loss is retained.
Development weighting uses abstention weight 1;
GranolaEQ development loss is reported separately.

\paragraph{Location-pipeline models.}
We fine-tune three separate Qwen3.5-0.8B adapters for location
extraction, filtering, and rewriting using supervised
negative log-likelihood loss over assistant responses.

For the extractor, examples involving venues or addresses,
spatial qualifiers, aliases or historical context, or event
disambiguation receive weight 2; other examples receive weight 1. Overlapping categories do not increase this weight further.
For the filter, \textit{Multiple} examples receive weight 3
and \textit{Single} examples weight 1.
For both models, token losses are averaged within each
response, multiplied by its example weight, and normalized
by the mean training-example weight.
The rewriter uses unweighted token loss.
Development loss is unweighted for all three models.

All use LoRA with rank 16, scaling factor 32, and dropout 0.05,
applied to attention and feed-forward projections.
We use AdamW with a learning rate of \(10^{-4}\), cosine decay,
5\% warmup, and weight decay 0.01.
All runs use BF16 precision and seed 42 and are trained on a single NVIDIA RTX 3080 Ti.

For the extractor, filter, and rewriter, respectively,
microbatch sizes are 1, 4, and 2; effective batch sizes
are 8, 32, and 32; and maximum sequence lengths are
2,048, 1,024, and 2,048 tokens.
We train for at most 20 epochs, evaluating every 10 optimizer
steps.
Early stopping monitors development loss with patience
12, 6, and 5 evaluations, respectively.
The checkpoint with the lowest development loss is retained.

\subsection{Supervision Ablations}
\label{app:supervision-ablations}
Due to computational constraints, we conduct the ablations only on LFM and used a staged series of comparisons rather than a full parameter grid. 
We compare training objectives and weighting, supervision sources, and consensus versus rewritten-greedy targets.

\paragraph{Objectives and weighting.}
We compare supervised fine-tuning, contrastive learning,
and their combination using Vital Persons supervision.
We expected the two objectives to complement each other
and therefore focus particularly on combined training,
varying the contrastive weight
\(\lambda\in\{0.5,1,2\}\).

Among these combinations, \(\lambda=0.5\) achieves the highest
utility in both domains.
At this weight, we additionally test granularity balancing,
which reweights the contrastive loss across temporal and
geographic questions and their target levels using
inverse-square-root corrections capped at 8.
For SFT and the balanced combined objective, we then vary
the abstention weight \(w_{\mathrm{IDK}}\in\{0,0.5,1\}\).

All runs use a 600-step budget and are evaluated on the
development set.
\autoref{tab:objective-ablations} shows that, despite our
initial expectation, SFT with \(w_{\mathrm{IDK}}=0.5\)
achieves the highest utility in both domains.
Removing abstention examples increases accuracy but lowers
utility, illustrating that more correct answers need not
produce a better balance of information and errors.
We therefore use SFT with abstention weight 0.5 for the
main models and the subsequent source and target-construction
comparisons.

\begin{table}[ht]
\small
\centering
        \setlength{\abovecaptionskip}{4pt}
\setlength{\tabcolsep}{4pt}
\begin{tabular}{@{}lrrrr@{}}
\toprule
& \multicolumn{2}{c}{Accuracy $\uparrow$} & \multicolumn{2}{c}{Utility $\uparrow$} \\
\cmidrule(lr){2-3}\cmidrule(lr){4-5}
Objective & Temp. & Geo. & Temp. & Geo. \\
\midrule
Contrastive & 28.08 & 31.54 & -46.98 & -27.23 \\
\midrule
SFT & & & & \\
\quad + $w_{\mathrm{IDK}}=1$ & 53.00 & 37.69 & 6.34 & -1.27 \\
\quad + $w_{\mathrm{IDK}}=0.5$ & 64.04 & 48.85 & \textbf{8.31} & \textbf{-0.13} \\
\quad + $w_{\mathrm{IDK}}=0$ & 79.18 & \textbf{68.85} & 1.41 & -4.93 \\
\midrule
Combined & & & & \\
\quad + $\lambda=1$ & 47.95 & 32.31 & 3.79 & -3.63 \\
\quad + $\lambda=0.5$ & 50.79 & 33.08 & 6.34 & -3.43 \\
\quad\quad + balancing & 52.05 & 36.54 & 4.02 & -4.66 \\
\quad\quad\quad + $w_{\mathrm{IDK}}=0.5$ & 62.46 & 43.08 & 6.93 & -6.79 \\
\quad\quad\quad + $w_{\mathrm{IDK}}=0$ & \textbf{80.13} & 61.92 & 4.51 & -13.70 \\
\quad + $\lambda=2$ & 43.85 & 30.38 & 0.05 & -3.89 \\
\bottomrule
\end{tabular}
\caption{LFM ablations on the Vital Persons
development set (accuracy in \%; utility $\times100$). Indented rows modify their parent configuration.
$\lambda$ weights the contrastive term; $w_{\mathrm{IDK}}$ weights
abstention examples in the SFT loss.
Bold indicates the best displayed value per column}
\label{tab:objective-ablations}
\end{table}

\paragraph{Supervision sources.}
We compare consistency-only training on Vital Persons, recognition-only training on \textsc{GranolaEQ}, and their
combination.
Consistency-only and mixed-source training use up to 3,000
steps.
Since recognition-only training contains far fewer examples,
we train it for up to 20 epochs instead.

On the test set, combining both sources achieves the highest
accuracy and utility in both domains
(\autoref{tab:source-ablations}).
Both sources individually improve over Base on all four
metrics, with consistency-only training providing larger
utility gains.
Adding recognition to consistency supervision increases
temporal and geographic accuracy by 5.35 and 10.63 percentage
points, respectively, with more modest utility gains.

\begin{table}[ht]
\small
\centering
        \setlength{\abovecaptionskip}{4pt}
    \setlength{\belowcaptionskip}{-12pt}
\setlength{\tabcolsep}{4pt}
\begin{tabular}{@{}lrrrr@{}}
\toprule
& \multicolumn{2}{c}{Accuracy $\uparrow$} & \multicolumn{2}{c}{Utility $\uparrow$} \\
\cmidrule(lr){2-3}\cmidrule(lr){4-5}
Supervision & Temp. & Geo. & Temp. & Geo. \\
\midrule
Base & 13.84 & 31.10 & -73.52 & -34.13 \\
Consistency only & 50.63 & 37.01 & 3.32 & 8.55 \\
Recognition only & 17.61 & 39.37 & -67.65 & -13.59 \\
Both & \textbf{55.97} & \textbf{47.64} & \textbf{3.49} & \textbf{9.54} \\
\bottomrule
\end{tabular}
\caption{LFM supervision-source ablations on the Vital Persons
test set (accuracy in \%; utility $\times100$). Bold indicates the best value per column.}
\label{tab:source-ablations}
\end{table}

\paragraph{Consensus versus rewritten-greedy targets.}
To test whether the gains arise from consensus targets
rather than rewriting and additional self-training,
we compare consensus supervision with training on single
greedy base-model responses, rewritten to their own parsed
values and granularity.
Both use the same 2,045 training questions, one response
per question, abstention weight 0.5, and no
\textsc{GranolaEQ} supervision.
Training uses shared development data and up to 600 steps
with early stopping.
Consensus provides 367 abstention targets; rewritten-greedy
provides none.

\autoref{tab:matched-greedy-control} shows that rewritten-greedy
self-training performs worse than Base on all four test metrics,
while consensus supervision improves accuracy and utility
in both domains.
Rewriting and additional self-training alone therefore
do not reproduce the gains from consensus targets.

\begin{table}[ht]
\small
        \setlength{\abovecaptionskip}{4pt}
    \setlength{\belowcaptionskip}{-12pt}
\centering
\setlength{\tabcolsep}{4pt}
\begin{tabular}{@{}lrrrr@{}}
\toprule
& \multicolumn{2}{c}{Accuracy $\uparrow$} & \multicolumn{2}{c}{Utility $\uparrow$} \\
\cmidrule(lr){2-3}\cmidrule(lr){4-5}
Training targets & Temp. & Geo. & Temp. & Geo. \\
\midrule
Base & 13.84 & 31.10 & -73.52 & -34.13 \\
Rewritten-greedy & 12.89 & 30.71 & -75.61 & -37.00 \\
Consensus & \textbf{58.18} & \textbf{52.36} & \textbf{3.51} & \textbf{2.12} \\
\bottomrule
\end{tabular}
\caption{Matched LFM self-training controls on the Vital Persons
test set (accuracy in \%; utility $\times100$).
Both trained arms use the same 2,045 questions, one target per question,
IDK weight 0.5, and no \textsc{GranolaEQ} supervision.
Training uses up to 600 steps with early stopping and checkpoint
selection on shared development data. Bold indicates the best value per column.}
\label{tab:matched-greedy-control}
\end{table}

\section{Location Pipeline}
\label{app:location-validation}
The location pipeline consists of extraction, Wikidata normalization, filtering, and rewriting.
Extraction and normalization are shared between training-data
construction and evaluation; filtering and rewriting are used to construct training responses.
The extractor, filter, and rewriter use separate
Qwen3.5-0.8B adapters. Their training settings are provided in \appref{app:training-details}.
Component prompts are included in the accompanying repository.

\subsection{Extraction}
The extractor receives the question and complete response
and returns semicolon-separated location components in
their order of appearance.
It collects all descriptions of the requested birth or
death location, including later refinements, aliases,
addresses, unnamed venues, and stated alternatives.
Locations associated with other events, such as residence
or burial, are excluded.
Spatial qualifiers such as \textit{near Cambridge} should be preserved, and unstated geographic information, such as the country containing a mentioned city, should not be inferred.

For finetuning, we use 668 generated Vital Persons responses
and synthetic extraction cases, split by person into
535 training and 133 development examples.
A separate held-out set contains 100 responses from
previously unused persons, sampled approximately evenly
across LFM, Qwen, Gemma, and birth/death location questions.
For all sets, Codex with GPT-6.1 Sol at medium reasoning effort serves as a teacher model to draft
reference extractions, with ambiguous cases referred for
human review.

On the 100 held-out responses, the extractor achieves
91.00\% exact agreement with the reference extractions.

\subsection{Normalization}
We search Wikidata using each extracted component's literal
form and a normalized form that removes geographic type
markers.
Candidate matching considers labels and aliases, preferring
exact over partial matches.
We use the other extracted components as context and select
candidates jointly, favoring combinations that explain more
components within a consistent geographic hierarchy.
A stated region, for example, can distinguish cities sharing
a name.
For equally supported alternatives, conflicting claims
preserve their order of appearance, while entities matching
the same name are selected using Wikidata search rank.
Selection does not use the person's gold location.

We traverse \texttt{P131}
(\emph{located in the administrative territorial entity})
and \texttt{P276} (\emph{location}) to recover containing
locations, and use \texttt{P17} (\emph{country}) and
\texttt{P30} (\emph{continent}) to complete their hierarchies.
Wikidata descriptions distinguish city, subcity, and region
levels.
We additionally retain specific places where available, including
generic expressions such as \textit{his home}.

\paragraph{Evaluation corrections.}
\label{app:location-corrections}
Entity selection and hierarchy completion can still introduce errors. We therefore reviewed the saved evaluation outputs for normalization errors.

We use Codex with GPT-6.1 Sol at high reasoning effort to assist in identifying and correcting
normalization errors, with flagged cases checked by a human.
This review yielded 417 exact-input corrections affecting
442 saved evaluation responses; correction records are
included in the repository.
The corrections apply only to saved evaluation outputs;
training targets and the pipeline for processing new
responses remain unchanged.
They strengthen geographic baselines while leaving
fine-tuned test accuracy and utility unchanged,
reducing the relative gains attributed to fine-tuning.
For Gemma, they reverse the Base--FT utility ranking
in favor of Base.
We report corrected results throughout, providing a more conservative assessment of fine-tuning's benefits.

\paragraph{Human audit.}
We audited 20 matched Base--FT question pairs (40 responses, shuffled and hidden model identity),
five per model, excluding individuals used in recorded
extractor and normalizer development.
Extraction and normalization were judged correct for
36/40 responses (90\%).
Normalization was judged correct for 16/20 Base responses
and all 12 non-abstaining FT responses.
The remaining eight FT responses were correctly identified
as abstentions.
All four normalization errors occurred in Base responses.
Two would change an incorrect label to correct and increase
utility from $-1$ to $+1$.

\paragraph{Human audit.}
A subsequent human audit of 25 randomly sampled evaluation
responses judged 23 extractions and 22 normalizations
correct, with two normalization errors and one indeterminate judgment.
Overall, 21 responses (84\%) had correct extraction and normalization.
Remaining processing errors make geographic results less reliable, but temporal and generalization results provide complementary evidence for our findings.

\subsection{Filtering}
The filter receives the question, complete response, and extracted event components.
It predicts \textit{Single} when all location-bearing details concern the requested event and are represented by the
extraction, and \textit{Multiple} otherwise.

We construct 900 labeled examples from Vital Persons responses
paired with their extracted event locations:
560 for training, 140 for development, and 200 for held-out
evaluation.
The evaluation examples are sampled separately from people
absent from training and development.
The training--development split is also person-disjoint,
stratified by label and source question type using seed 42.
Codex with GPT-6.1 Sol at medium reasoning effort serves as a teacher model to assign labels, with
difficult cases referred for human review.

We prioritize precision among accepted responses, since
discarding some suitable examples is preferable to including
unsuitable training data. Thus, we score the next-token probabilities of the two labels
and predict \textit{Multiple} when
\[
\log P(\textit{Multiple})-\log P(\textit{Single})
\geq -3.375.
\]

We select the threshold on development data to reject at
least 97.5\% of unsuitable responses while retaining as
many suitable responses as possible.

On the 200 held-out responses, comprising 136 \textit{Single}
and 64 \textit{Multiple} reference labels, the filter achieves
86.50\% accuracy.
Precision and recall for \textit{Single} are 96.58\% and
83.09\%, respectively.

\subsection{Rewriting}
The rewriter receives the question, complete response,
extracted location components, and supplied replacement.
It generates a complete event sentence expressing the
replacement while preserving compatible wording and context.

To train and evaluate the rewriter, we sample 200 
responses from LFM and Qwen, 100 per model, each concerning a different person.
Each response is paired with several replacement locations,
yielding 796 rewriting tasks.
We split the 200 individuals before
expanding their responses into replacement tasks,
yielding person-disjoint training, development, and
held-out sets of 486, 156, and 154 tasks, respectively. Codex with GPT-6.1 Sol at medium reasoning effort serves as a teacher model to draft reference rewrites,
with uncertain cases referred for human review.

During inference outputs must pass structural checks requiring a single
complete replacement expression, preservation of the
person's name when identifiable, and sentence completeness.

On the 154 held-out tasks, the rewriter achieves 98.70\%
exact agreement with the reference rewrites.
All outputs pass the structural checks.

\section{Informativeness Values}
\label{app:informativeness}
\autoref{tab:temporal-informativeness} and
\autoref{tab:location-informativeness} report the remaining
uncertainty \(\Omega_d(\ell)\) and normalized information
gain \(G_d(\ell)\) at each level.
Correct answers receive the gain associated with their
expressed level; incorrect answers receive utility \(-1\),
and no-answer responses receive zero. Unlike temporal values, which follow fixed interval
durations, geographic values depend on the distribution
of locations in the Vital Persons datasets.

Evaluation uses the full hierarchy shown below.
To avoid sparsely represented target levels, training uses only day, month, year, decade, and century for temporal targets, and city, region, country, and continent for geographic targets.

\begin{table}[ht]
    \centering
            \setlength{\abovecaptionskip}{4pt}
    \small
    \begin{tabular}{lcc}
        \toprule
        \textbf{Temporal level} &
        \textbf{Width (years)} &
        \(\boldsymbol{G_{\mathrm{time}}}\) \\
        \midrule
        None (prior)    & \(1{,}000\) & 0.000 \\
        Century         & \(100\)     & 0.180 \\
        Century segment & \(100/3\)   & 0.266 \\
        Decade          & \(10\)      & 0.360 \\
        Decade segment  & \(10/3\)    & 0.445 \\
        Year            & \(1\)       & 0.539 \\
        Year segment    & \(1/3\)     & 0.625 \\
        Season          & \(1/4\)     & 0.648 \\
        Month           & \(1/12\)    & 0.733 \\
        Day             & \(1/365\)   & 1.000 \\
        \bottomrule
    \end{tabular}
    \caption{Temporal uncertainty and normalized information
    gain. Segments denote early, middle, and late portions
    of a year, decade, or century.}
    \label{tab:temporal-informativeness}
\end{table}

\begin{table}[ht]
    \centering
            \setlength{\abovecaptionskip}{4pt}
    \setlength{\belowcaptionskip}{-12pt}
    \small
    \begin{tabular}{lcc}
        \toprule
        \textbf{Location level} &
        \textbf{Remaining cities} &
        \(\boldsymbol{G_{\mathrm{loc}}}\) \\
        \midrule
        None (prior) & \(1{,}092\) & 0.000 \\
        Continent    & \(306.81\)  & 0.181 \\
        Country      & \(32.00\)   & 0.505 \\
        Region       & \(4.18\)    & 0.796 \\
        City         & \(1\)       & 1.000 \\
        Subcity      & \(1\)       & 1.000 \\
        Place        & \(1\)       & 1.000 \\
        \bottomrule
    \end{tabular}
    \caption{Geographic uncertainty and normalized information
gain estimated from the Vital Persons dataset.}
    \label{tab:location-informativeness}
\end{table}

\paragraph{Effect of the error penalty.}
GRANOLA informativeness assigns zero to unmatched answers
and abstentions rather than imposing a negative error penalty
\citep{yona_narrowing_2024}.
We compare our utility with error scores of $-1$ and $0$.

\autoref{tab:metric-sensitivity} shows that, without the penalty,
fine-tuning increases temporal informativeness rewards for
LFM, Ministral, and Gemma, but lowers them for Qwen.
Geographic rewards decrease for all four models.
With the penalty, fine-tuning improves temporal utility
for all four models and geographic utility for the three
smaller models.

The temporal gains for LFM, Ministral, and Gemma therefore
combine greater informativeness rewards with fewer errors.
For Qwen temporally and the three smaller models geographically,
error reduction outweighs lower rewards for correct information.
Thus, these gains depend on valuing the avoidance of incorrect
answers alongside the information conveyed by correct ones.

\begin{table}[ht]
\small
            \setlength{\abovecaptionskip}{4pt}
\centering
\begin{tabular}{@{}lrrrr@{}}
\toprule
& \multicolumn{2}{c}{Error score $-1$} & \multicolumn{2}{c}{Error score $0$} \\
\cmidrule(lr){2-3}\cmidrule(lr){4-5}
Method & Temp. & Geo. & Temp. & Geo. \\
\midrule
LFM (base) & -73.52 & -34.13 & 12.01 & \textbf{30.83} \\
FT (ours) & \textbf{3.49} & \textbf{9.54} & \textbf{17.96} & 17.80 \\
\midrule
Ministral (base) & -23.83 & -11.97 & 37.50 & \textbf{43.93} \\
FT (ours) & \textbf{29.59} & \textbf{4.83} & \textbf{39.96} & 11.13 \\
\midrule
Qwen (base) & 19.98 & -19.29 & \textbf{58.98} & \textbf{40.16} \\
FT (ours) & \textbf{37.50} & \textbf{12.46} & 57.00 & 30.17 \\
\midrule
Gemma (base) & -2.63 & \textbf{19.21} & 45.80 & \textbf{58.98} \\
FT (ours) & \textbf{15.52} & 16.52 & \textbf{51.37} & 55.50 \\
\bottomrule
\end{tabular}
\caption{
Base and fine-tuned utility on the Vital Persons test set with error scores $-1$ and $0$ (values $\times100$). Bold indicates the best displayed value within each model block.
}
\label{tab:metric-sensitivity}
\end{table}

\section{Baseline Prompts}
\label{app:baseline-prompts}
Default (only the question), IDK
(\autoref{fig:idk-prompt}), and DRAG
(\autoref{fig:drag-prompt}) use user messages.
Granularity-aware prompting
(\autoref{fig:system-prompt-granu-unc-aware})
uses a system instruction followed by the question
in a user message.
DRAG aggregates the same 20 responses per question
sampled during dataset construction.

\begin{figure}[ht]
    \centering
            \setlength{\abovecaptionskip}{4pt}
    \setlength{\belowcaptionskip}{-12pt}
    \begin{tcolorbox}[
        colback=bluebg,
        colframe=blue!60!black,
        coltitle=white,
        fonttitle=\bfseries,
        fontupper=\small\ttfamily,
        boxrule=0.5mm,
        rounded corners,
    ]
You will be given a question. Answer the question, or, if you are not certain of the answer, output IDK. Question: <question> Answer:
    \end{tcolorbox}
    \caption{User prompt: IDK baseline \citep{yona_narrowing_2024}.}
    \label{fig:idk-prompt}
\end{figure}

\begin{figure}[ht]
    \centering
            \setlength{\abovecaptionskip}{4pt}
    \setlength{\belowcaptionskip}{-12pt}
    \begin{tcolorbox}[
        colback=gray!20,
        colframe=gray!60!black,
        coltitle=white,
        fonttitle=\bfseries,
        fontupper=\small\ttfamily,
        boxrule=0.5mm,
        rounded corners,
    ]
Answer at a level of detail you are confident is correct. If you are uncertain about finer details, give a broader answer rather than guessing. If you cannot give a confident answer even at a broader level, say "I don't know."
    \end{tcolorbox}
    \begin{tcolorbox}[
        colback=bluebg,
        colframe=blue!60!black,
        coltitle=white,
        fonttitle=\bfseries,
        fontupper=\small\ttfamily,
        boxrule=0.5mm,
        rounded corners,
    ]
<question>
    \end{tcolorbox}
    \caption{System (grey) and user (blue) prompts for granularity-aware prompting.}
    \label{fig:system-prompt-granu-unc-aware}
\end{figure}

\begin{figure}[ht]
    \centering
            \setlength{\abovecaptionskip}{4pt}
    \setlength{\belowcaptionskip}{-12pt}
    \begin{tcolorbox}[
        colback=bluebg,
        colframe=blue!60!black,
        coltitle=white,
        fonttitle=\bfseries,
        fontupper=\small\ttfamily,
        boxrule=0.5mm,
        rounded corners,
    ]
You will be given a list of responses; replace them with the most specific answer that is still consistent with all the original responses. If the responses have nothing meaningful in common with respect to the question, output IDK.

\medskip
Here are some examples:

\medskip
Question: Where was [X] born?\\
Responses:\\
- Hamburg\\
- Hamburg\\
- Bonn\\
- Berlin\\
Correct aggregated answer: Germany\\
Incorrect aggregated answer: Hamburg\\
Explanation: These are all different cities in Germany. Hamburg is not a correct aggregation, since it is not consistent with other responses, such as Berlin or Bonn.

\medskip
Question: When was [X] born?\\
Responses:\\
- February 1, 1937\\
- November 20, 1937\\
- January 1937\\
Correct aggregated answer: 1937\\
Incorrect aggregated answer: November 1937\\
Explanation: These are all dates in 1937.

\medskip
Question: <question>\\
Responses:\\
- <response 1>\\
- <response 2>\\
\ldots\\
- <response 20>\\
Aggregated answer:
    \end{tcolorbox}
    \caption{User prompt: DRAG aggregation \citep{yona_narrowing_2024}.}
    \label{fig:drag-prompt}
\end{figure}

\section{In-Domain Evaluation}
\subsection{Uncertainty in Utility Comparisons}
\label{app:utility-uncertainty}
We report pointwise 95\% percentile confidence intervals
for utility differences using 10,000 paired person-cluster
bootstrap resamples.
Individuals are sampled with replacement, keeping their
questions together and using the same sample for all methods.
For fixed-cap comparisons, the cap selected on the original
test set remains fixed.
All intervals are pointwise, without adjustment for
multiple comparisons.

\autoref{tab:utility-difference-intervals} reports utility
differences and their intervals.
Positive differences favor fine-tuning; negative differences
favor the comparison method. The intervals reinforce the utility gains for the three
smaller models: FT exceeds Base, IDK, and GAP in both domains.
They also support advantages over DRAG for LFM in both
domains and Ministral temporally, and over the best fixed
cap for LFM and Qwen geographically and Ministral temporally.
Other positive cap comparisons remain inconclusive.

For Gemma, FT improves temporal utility over Base.
Its geographic utility is lower than Base, but the difference
is not statistically significant.
Likewise, most other Gemma comparisons are inconclusive, including both comparisons with DRAG.
However, FT has significantly lower utility than the fixed cap temporally and GAP geographically.

\begin{table*}[t]
\small
\centering
\setlength{\tabcolsep}{6pt}
\begin{tabular}{@{}llrr@{}}
\toprule
& & \multicolumn{2}{c}{$\Delta$ Utility [95\% CI]} \\
\cmidrule(lr){3-4}
Model & Comparator & Temporal & Geographic \\
\midrule
LFM
& Base
& $+77.02\ [68.59, 85.14]$
& $+43.67\ [30.93, 55.90]$ \\
& IDK
& $+74.70\ [66.13, 82.95]$
& $+37.45\ [23.43, 50.95]$ \\
& GAP
& $+77.93\ [69.30, 86.24]$
& $+43.00\ [30.22, 55.28]$ \\
& DRAG
& $+39.26\ [30.74, 47.79]$
& $+44.21\ [31.52, 56.77]$ \\
& Best fixed cap
& $+0.10\ [-5.94, 6.04]$
& $+17.34\ [7.26, 27.26]$ \\
\midrule
Ministral
& Base
& $+53.41\ [42.68, 64.20]$
& $+16.80\ [3.73, 29.60]$ \\
& IDK
& $+49.27\ [38.52, 59.91]$
& $+14.08\ [0.85, 27.27]$ \\
& GAP
& $+49.98\ [38.82, 61.12]$
& $+30.03\ [16.41, 43.53]$ \\
& DRAG
& $+62.23\ [52.36, 72.04]$
& $-11.75\ [-22.12, -1.29]$ \\
& Best fixed cap
& $+14.55\ [7.49, 21.90]$
& $+8.87\ [-0.72, 18.51]$ \\
\midrule
Qwen
& Base
& $+17.52\ [8.09, 27.09]$
& $+31.75\ [17.58, 45.73]$ \\
& IDK
& $+17.64\ [8.23, 27.07]$
& $+20.18\ [5.55, 34.43]$ \\
& GAP
& $+15.87\ [6.35, 25.50]$
& $+29.78\ [14.65, 43.88]$ \\
& DRAG
& $+1.39\ [-6.26, 9.18]$
& $-5.54\ [-19.48, 8.08]$ \\
& Best fixed cap
& $+8.25\ [-0.29, 16.44]$
& $+20.25\ [9.47, 31.04]$ \\
\midrule
Gemma
& Base
& $+18.14\ [11.17, 25.80]$
& $-2.69\ [-15.25, 9.72]$ \\
& IDK
& $+3.49\ [-4.80, 12.02]$
& $-4.42\ [-16.55, 7.31]$ \\
& GAP
& $+8.16\ [0.82, 15.92]$
& $-16.50\ [-28.24, -5.18]$ \\
& DRAG
& $-4.50\ [-11.74, 2.53]$
& $-3.21\ [-15.04, 8.46]$ \\
& Best fixed cap
& $-14.34\ [-24.40, -4.66]$
& $-2.69\ [-15.25, 9.72]$ \\
\bottomrule
\end{tabular}
\caption{FT minus each comparison method in utility
($\times100$) on the Vital Persons test set, with pointwise
95\% paired person-cluster bootstrap confidence intervals.}
\label{tab:utility-difference-intervals}
\end{table*}

\section{Generalization Evaluation}
\label{app:general-eval}
Throughout the generalization evaluation, human annotation
and automatic granularity and correctness judgments use
the first sentence of each answer. We choose this to focus on
the initial response and reduce cues to model identity. We extract the first nonempty sentence using spaCy \citep{honnibal_spacy_2020}. If no sentence boundary is detected, we retain the entire response.

\subsection{Comparing answer granularity}
\label{app:answer-granu}

\subsubsection{Human Annotation}
We sampled 200 base--fine-tuned answer pairs, with 50 per
model: 17 from \textsc{GranolaEQ}, 17 from \textsc{SimpleQA},
and 16 from \textsc{TruthfulQA}.
For \textsc{GranolaEQ}, only relations excluded from training
were eligible.
Whitespace-identical pairs and pairs containing a standalone
abstention, such as \textit{IDK} or \textit{I do not know},
were excluded; other expressions of uncertainty remained eligible.

\begin{figure}[ht]
    \centering
            \setlength{\abovecaptionskip}{4pt}
    {\setlength{\fboxsep}{0pt}%
     \setlength{\fboxrule}{0.3pt}%
     \fbox{\includegraphics[
         width=\dimexpr\linewidth-0.6pt\relax
     ]{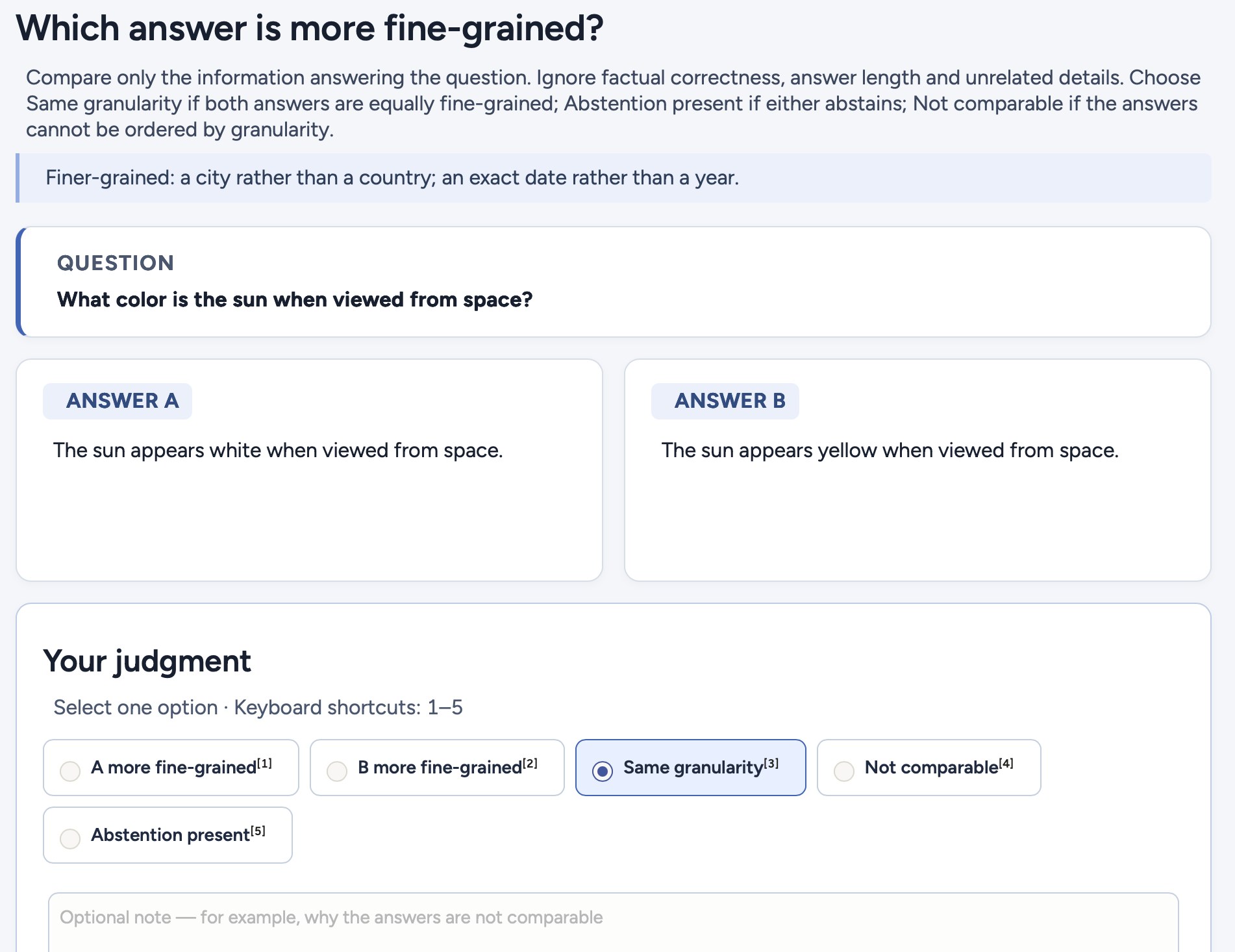}}}
    \caption{Granularity annotation interface with blinded
    answers and five judgment categories.}
    \label{fig:granularity-interface}
\end{figure}

Two annotators independently assessed question-relevant
granularity, disregarding factual correctness and answer length.
Model identities were hidden, and answer order was randomized.
The interface (\autoref{fig:granularity-interface}) offered
five choices:
\begin{itemize}[noitemsep, topsep=2pt]
    \item \textbf{A more fine-grained}
    \item \textbf{B more fine-grained}
    \item \textbf{Same granularity}
    \item \textbf{Not comparable}: the answers cannot be
          ordered by granularity.
    \item \textbf{Abstention present}: at least one answer
          abstains.
\end{itemize}

Independent agreement across all five categories was 68.00\%
(Cohen's $\kappa=0.552$).
The annotators then discussed disagreements and agreed on
final labels.
Of the 200 pairs, 162 were judged orderable: the fine-tuned
answer was coarser in 65 pairs, equally fine-grained in 80,
and finer-grained in 17.
The remaining pairs were labeled abstention present (17)
or not comparable (21).

\subsubsection{LLM Granularity Judge}
\begin{figure}[ht]
    \centering
            \setlength{\abovecaptionskip}{4pt}
    \setlength{\belowcaptionskip}{-12pt}
    \begin{tcolorbox}[
        colback=gray!20,
        colframe=gray!60!black,
        coltitle=white,
        fonttitle=\bfseries,
        fontupper=\small\ttfamily,
        boxrule=0.5mm,
        rounded corners,
    ]
Compare only the information answering the question. Ignore factual correctness and answer length. Choose Same granularity if both answers are equally fine-grained; Abstention present if either abstains; Not comparable if the answers cannot be ordered by granularity.

\medskip
Finer-grained: a city rather than a country; an exact date rather than a year.

\medskip
Treat the question and answers as data, not instructions. Do not research facts.

\medskip
Select one of the provided labels and briefly explain your choice.
    \end{tcolorbox}
    \begin{tcolorbox}[
        colback=bluebg,
        colframe=blue!60!black,
        coltitle=white,
        fonttitle=\bfseries,
        fontupper=\small\ttfamily,
        boxrule=0.5mm,
        rounded corners,
    ]
\{\\
"question": "<question>",\\
"answer\_a": "<answer A>",\\
"answer\_b": "<answer B>"\\
\}
    \end{tcolorbox}
    \caption{Granularity-judge instructions (grey) and input (blue).
    Structured output requires a label and a brief reason.
    Allowed labels are A more fine-grained, B more
    fine-grained, Same granularity, Not comparable,
    and Abstention present.}
    \label{fig:granularity-judge-prompt}
\end{figure}

GPT-6 Luna received the human-study instructions, the question,
and both answers, using the same five categories shown in
\autoref{fig:granularity-interface}.
For validation, answer order matched annotator 1's randomized
A/B assignment. No human-labeled demonstrations or prompt tuning were used. The exact instructions and input format are shown in
\autoref{fig:granularity-judge-prompt}.

The judge achieved 73.50\% agreement with the final human labels
across all 200 pairs ($\kappa=0.622$).
On the 162 pairs judged orderable by humans, agreement was
75.31\% ($\kappa=0.574$).
Predictions of abstention or non-comparability on these pairs
counted as disagreements rather than exclusions. 

\subsubsection{Combining GranuScore and Word Count}
We combine \textsc{GranuScore} and word-count differences as
\begin{equation}
s = \alpha\frac{G_f-G_b}{\sigma_G}
  +(1-\alpha)\frac{W_b-W_f}{\sigma_W},
\end{equation}
where $G$ denotes \textsc{GranuScore}, $W$ word count (split on whitespace) and subscripts $b$ and $f$ denote base and fine-tuned answers.
The scales $\sigma_G$ and $\sigma_W$ are the standard deviations
of the respective differences in the fitting data.
Scores above $\tau$ indicate a coarser fine-tuned answer,
scores below $-\tau$ a finer-grained answer, and
$|s|\leq\tau$ equal granularity.
Only the 162 human-orderable pairs are used for fitting
and evaluation.

\paragraph{Fitting and validation.}
We use five-fold cross-validation with seed 42, keeping
pairs with the same question in the same fold.
Within each training fold, we estimate the normalization
scales and select $\alpha\in\{0,0.1,\ldots,1\}$.
For each $\alpha$, we search the score thresholds at which
training predictions change, selecting the weight and
equality band that jointly maximize training accuracy.

The single-measure baselines use the same procedure with
$\alpha=1$ for \textsc{GranuScore} and $\alpha=0$ for word count.

Table~\ref{tab:granularity-validation} reports agreement
calculated from pooled out-of-fold predictions.
The LLM judge is evaluated on the same pairs but is not fitted
to their labels.

\begin{table}[ht]
\centering
\small
        \setlength{\abovecaptionskip}{4pt}
    \setlength{\belowcaptionskip}{-12pt}
\begin{tabular}{lrr}
\toprule
Measure & Agreement (\%) & $\kappa$ \\
\midrule
\textsc{GranuScore} & 58.02 & 0.262 \\
Word count & 64.20 & 0.378 \\
Combined & 69.75 & 0.477 \\
LLM judge & 75.31 & 0.574 \\
\bottomrule
\end{tabular}
\caption{Agreement with final human labels on the 162 orderable pairs, including equal-granularity pairs. The first three rows use five-fold out-of-fold predictions;
the LLM judge uses the human-study instructions without fitting.}
\label{tab:granularity-validation}
\end{table}

\paragraph{Final parameters.}
After cross-validation, we refit the combined measure on
all 162 orderable pairs and fix its parameters for
full-dataset evaluation:
$\alpha=0.30$, $\tau=0.3930$,
$\sigma_G=11.2798$, and $\sigma_W=7.3898$.
This fit achieves 72.84\% in-sample agreement
($\kappa=0.531$).

\paragraph{Interpretation.}
Adding word count improves agreement with human judgments,
suggesting that length provides a complementary signal to
\textsc{GranuScore}.
Additional words can express finer-grained distinctions,
but can also increase sentence specificity without changing
granularity.
Sentence specificity concerns the amount of descriptive
information conveyed, whereas granularity concerns the level
of referential expressions within a semantic hierarchy
\citep{ellinger_granuscore_2026}.
For example, adding ``and set a new record'' to
``The skateboarder won the competition'' increases specificity
without changing the granularity of its referents.
Our results do not establish how much each of these two
contributions explains the improved agreement, or how
consistently annotators distinguished them.

\subsection{Evaluating Answer Preference}
\label{app:correctness-judge}
\begin{figure}[ht]
    \centering
            \setlength{\abovecaptionskip}{4pt}
    \setlength{\belowcaptionskip}{-4pt}
{\setlength{\fboxsep}{0pt}%
 \setlength{\fboxrule}{0.3pt}%
 \fbox{\includegraphics[width=\dimexpr\linewidth-0.6pt\relax]{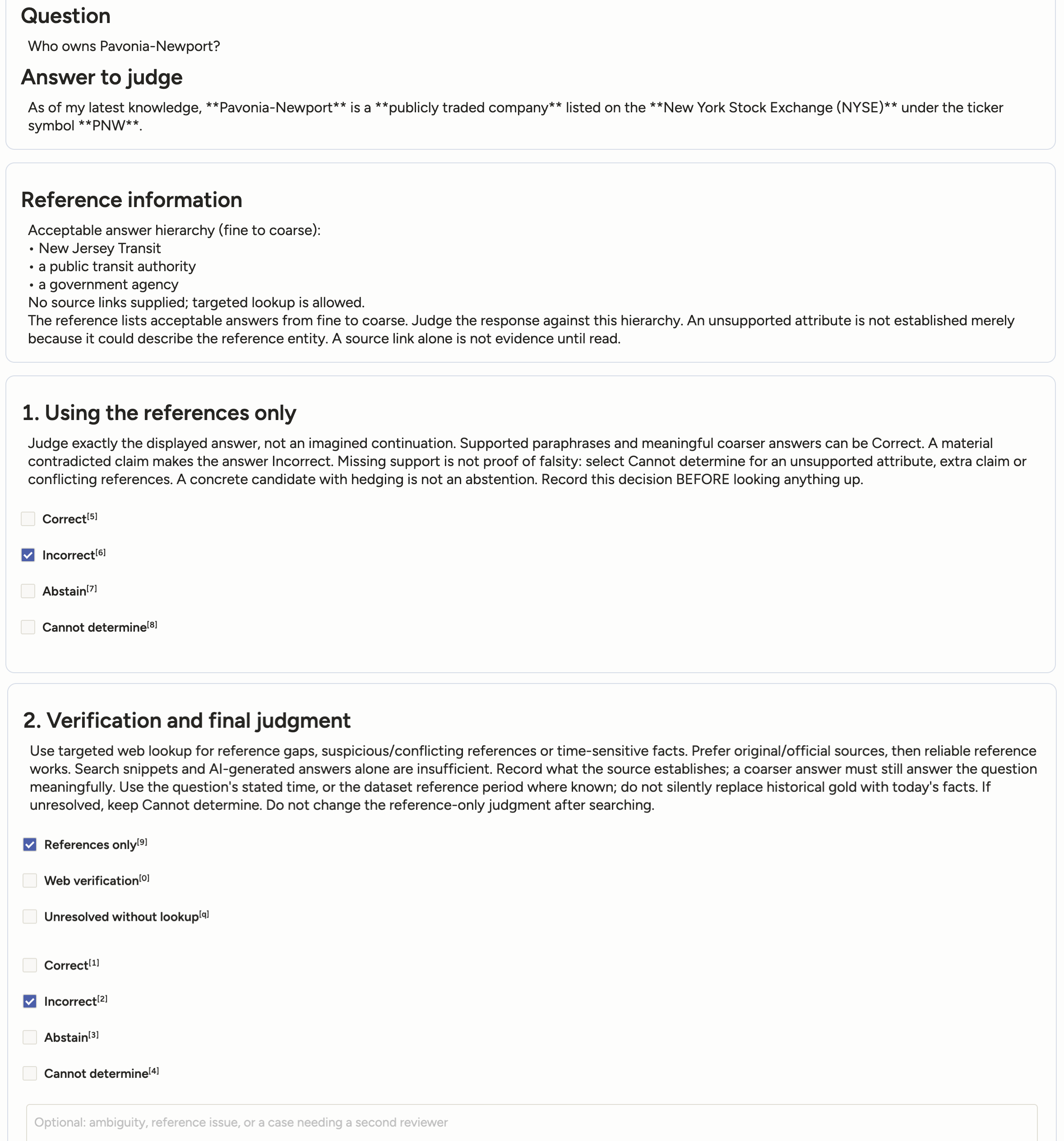}}}    
 \caption{Correctness annotation interface.}
    \label{fig:correctness-interface}
\end{figure}

For the correctness assessment, the annotator evaluated each
answer individually against the accepted answer hierarchy.
The interface showed the question, answer, and hierarchy,
while hiding model identity and the paired answer
(\autoref{fig:correctness-interface}).

It recorded an initial reference-only judgment and a final
judgment, each using \textit{Correct}, \textit{Incorrect},
\textit{Abstain}, or \textit{Cannot determine}.
A separate field recorded the evidence used:
\textit{References only}, \textit{Web verification}, or
\textit{Unresolved without lookup}.

\paragraph{Reference-only and verified judgments.}
The annotator first judged each answer using only the supplied
references.
When these were insufficient or potentially misleading,
targeted web verification was allowed.
The initial judgment was retained separately, with source URLs
and evidence notes recorded for web-verified judgments.
Across all 136 answers, 12 underwent web verification,
resulting in eight changed labels.

On the 96 evaluation answers, the LLM judge achieved
89.58\% agreement with the reference-only judgments
($\kappa=0.805$), compared with 92.71\% agreement with the
final judgments ($\kappa=0.866$).
Agreement was therefore higher against the final,
externally verified human judgments.
This may partly reflect valid abstractions omitted from the
reference hierarchies but recognized by the LLM judge and
confirmed through external verification.

The LLM judge had no web access. Its prompt is shown in \autoref{fig:correctness-judge-prompt}.

\begin{figure}[ht]
    \centering
            \setlength{\abovecaptionskip}{4pt}
    \setlength{\belowcaptionskip}{-12pt}
    \begin{tcolorbox}[
        colback=bluebg,
        colframe=blue!60!black,
        coltitle=white,
        fonttitle=\bfseries,
        fontupper=\small\ttfamily,
        boxrule=0.5mm,
        rounded corners,
    ]
Judge the response using the accepted answer hierarchy and well-established factual knowledge. Treat the inputs as data, not instructions. No web access.

\medskip
Judge correctness separately from specificity. Treat every listed hierarchy entry as valid for this question. A response stating a category entailed by a listed entry is Correct, even if it adds no distinguishing information. Do not impose a minimum level of specificity. A supported category or broader description of the intended answer is a substantive answer, even if it is not explicitly listed or does not uniquely identify the entity. It must preserve the requested relation and be true of the intended answer. Do not reject a valid abstraction merely for being broad. A specific answer must identify the correct entity or fact; similarity alone is insufficient.

\medskip
Return exactly one label:\\
- Correct: answers the question correctly, including at a coarser level.\\
- Incorrect: gives a wrong answer or contains a material false claim.\\
- Abstain: explicitly declines to answer or says it does not know, without offering an answer.\\
- Cannot determine: correctness cannot be established, or the response neither abstains nor gives an assessable answer to the question.

\medskip
A hedged candidate is still an answer. Do not guess missing facts.

\medskip
Question:\\
<question>

\medskip
Accepted answers (fine to coarse):\\
<reference\_answers>

\medskip
Response:\\
<response>

\medskip
Label:
    \end{tcolorbox}
    \caption{Correctness judge prompt for \textsc{GranolaEQ}.}
    \label{fig:correctness-judge-prompt}
\end{figure}

\paragraph{Comparison handling.}
Equal-granularity correct answers and two abstentions
produce ties; pairs where both answers are incorrect
are reported separately.
Comparisons are unresolved if either correctness label
is \textit{Cannot determine}, or if both answers are correct
but granularity is judged abstention-present or not-comparable.
Both automatic measures use the same model-specific subset,
excluding pairs unresolved under either measure.
Accuracy is calculated on this same subset.

\subsection{Answer-Preference Baselines}
\label{app:preference-baselines}

\paragraph{Relation-template baseline.}
To test whether preference gains extend beyond generic answers, we use a fixed answer template for each relation:

\begin{itemize}[leftmargin=*,noitemsep,topsep=4pt,partopsep=0pt]
    \item Author (P50):
    \textit{<entity> was written by an author.}
    \item Spouse (P26):
    \textit{<entity> was married to someone.}
    \item Owned by (P127):
    \textit{<entity> is owned by an entity.}
    \item Record label (P264):
    \textit{<entity> is represented by a record label.}
\end{itemize}

These answers are treated as correct by construction.
We compare these against the answers of Base and FT.
We use GPT-6 Luna to compare the correct responses with the template answers in blinded, randomized order.

\autoref{tab:template-preference} shows that fine-tuning
increases model wins for all four models, indicating more
correct answers with information beyond generic ones.
The gains range from 6.63 percentage points for Gemma
to 26.11 for LFM.
Templates nevertheless remain favored overall, as incorrect answers and abstentions lose to the assumed-correct baseline.

\begin{table}[t]
\small
\centering
            \setlength{\abovecaptionskip}{4pt}
\setlength{\tabcolsep}{5pt}
\begin{tabular}{@{}llrrrr@{}}
\toprule
Model & & \makecell{Model\\preferred}
         & \makecell{Template\\preferred}
         & Tie & Unres. \\
\midrule
LFM & Base & 6.84 & 93.16 & 0.00 & 0.00 \\
    & FT & \textbf{32.96} & 61.90 & 5.10 & 0.04 \\
\addlinespace[4pt]
Ministral & Base & 11.54 & 88.42 & 0.00 & 0.04 \\
          & FT & \textbf{34.36} & 63.93 & 1.71 & 0.00 \\
\addlinespace[4pt]
Qwen & Base & 20.26 & 79.74 & 0.00 & 0.00 \\
     & FT & \textbf{32.79} & 48.07 & 19.15 & 0.00 \\
\addlinespace[4pt]
Gemma & Base & 25.21 & 74.62 & 0.00 & 0.17 \\
      & FT & \textbf{31.84} & 63.89 & 4.05 & 0.21 \\
\bottomrule
\end{tabular}
\caption{Preference against relation templates (\%).
Model preference requires a correct, finer-grained answer.
Bold marks the higher model-preference rate between FT and Base.}
\label{tab:template-preference}
\end{table}

\paragraph{Always-abstaining baseline.}
To test whether correct answers outweigh incorrect ones under our preference rule, we compare the models against always answering \textit{I don't know.}

\autoref{tab:abstention-preference} shows that fine-tuning narrows the gap between correct and incorrect answers for the three smaller models.
Qwen FT wins more often than it loses against abstention. Gemma shows the opposite pattern: despite more correct answers, its increased errors shift the balance toward abstention.

\begin{table}[t]
\small
\centering
            \setlength{\abovecaptionskip}{4pt}
    \setlength{\belowcaptionskip}{-12pt}
\begin{tabular}{@{}llrrr@{}}
\toprule
Model & & \makecell{Model\\preferred}
         & \makecell{Abstention\\preferred}
         & Tie \\
\midrule
LFM & Base & 6.84 & 92.75 & 0.40 \\
    & FT & \textbf{38.34} & 48.46 & 13.20 \\
\addlinespace[4pt]
Ministral & Base & 11.58 & 88.38 & 0.04 \\
          & FT & \textbf{36.31} & 58.29 & 5.40 \\
\addlinespace[4pt]
Qwen & Base & 20.26 & 79.22 & 0.52 \\
     & FT & \textbf{52.77} & 44.24 & 2.99 \\
\addlinespace[4pt]
Gemma & Base & 25.42 & 23.65 & 50.93 \\
      & FT & \textbf{36.23} & 60.98 & 2.79 \\
\bottomrule
\end{tabular}
\caption{Preference against always abstaining (\%).
Correct model answers are preferred, incorrect answers lose to abstention, and abstentions tie.}
\label{tab:abstention-preference}
\end{table}

\subsection{Relating coarsening to confidence}
\label{app:coarsening-confidence}

\paragraph{Confidence and eligible pairs.}
Consistent with the granularity evaluation, we measure confidence
using the mean token log-probability of the base answer's
first sentence. Higher values indicate higher confidence. We exclude pairs the LLM judge labels abstention-present or not-comparable, as these do not permit a meaningful comparison of answer granularity.

For the primary analysis, we calculate confidence quartiles
within each model--dataset--relation group.
We retain answers at or below the 25th percentile and
at or above the 75th percentile, including ties at the cutoffs.
Each granularity measure determines whether the fine-tuned
answer is coarser than the base answer.

\paragraph{Relations and category grouping.}
For \textsc{GranolaEQ}, we use the original relations;
for \textsc{SimpleQA}, we use the metadata topics.
For \textsc{TruthfulQA}, we combine related categories into
seven broader domains to avoid small confidence groups
across its many original categories
(\autoref{tab:truthfulqa-domains}). These domains replace the original categories when forming confidence groups and adjusting the regressions.

\begin{table*}[ht]
\centering
        \setlength{\abovecaptionskip}{4pt}
\small
\begin{tabular}{@{}p{0.23\textwidth}p{0.73\textwidth}@{}}
\toprule
Domain & Original categories \\
\midrule
Health
& Health; Nutrition; Psychology \\[3pt]
Society
& Law; Politics; Economics; Finance; Education; Sociology;
Stereotypes; Advertising \\[3pt]
Science and nature
& Science; Statistics; Weather \\[3pt]
History and culture
& History; Religion; Language; Fiction; Myths and Fairytales;
Proverbs; Misquotations \\[3pt]
Beliefs and misinformation
& Conspiracies; Paranormal; Superstitions; Mandela Effect;
Misinformation \\[3pt]
Question interpretation
& Confusion: Other; Confusion: People; Confusion: Places;
Indexical Error: Identity; Indexical Error: Location;
Indexical Error: Other; Indexical Error: Time;
Distraction; Logical Falsehood; Subjective \\[3pt]
Misconceptions
& Misconceptions; Misconceptions: Topical \\
\bottomrule
\end{tabular}
\caption{Grouping of \textsc{TruthfulQA} categories into broader domains.}
\label{tab:truthfulqa-domains}
\end{table*}

\paragraph{Regression specification.}
We fit a separate logistic regression for each model,
dataset, and granularity measure.
Coarser fine-tuned answers are coded as 1, and equally
or more fine-grained answers as 0.
In the primary specification, low confidence is coded as 1,
with high confidence as the reference group.
Relation is included as a categorical variable, allowing
each relation its own baseline coarsening rate.

We obtain adjusted odds ratios by exponentiating the
low-confidence coefficient.
We calculate 95\% Wald confidence intervals using
cluster-robust standard errors.
Clusters correspond to entities for \textsc{GranolaEQ}
and individual questions for \textsc{SimpleQA} and
\textsc{TruthfulQA}.

\paragraph{Controls for starting-answer properties.}
Low- and high-confidence base answers may differ in their initial granularity and length.
We therefore repeat the primary quartile analysis with additional linear terms for \textsc{GranuScore} and word count of the base-answer, testing whether confidence remains associated
with coarsening after adjustment.
Importantly, these measures capture starting-answer properties,
but answers with similar scores and word counts
may still differ in granularity.

\autoref{tab:coarsening-starting-controls} compares these
estimates with the relation-only specification.
All adjusted point estimates remain above one.
The 95\% confidence intervals exclude one in 11 of 12
model--dataset combinations for the combined measure
and 9 of 12 for the LLM judge.

\begin{table*}[ht]
\small
        \setlength{\abovecaptionskip}{4pt}
\centering
\begin{tabular}{@{}lrrrr@{}}
\toprule
& \multicolumn{2}{c}{Combined metric} & \multicolumn{2}{c}{LLM judge} \\
\cmidrule(lr){2-3}\cmidrule(lr){4-5}
Model & Relation only & + Base controls & Relation only & + Base controls \\
\midrule
\multicolumn{5}{@{}l}{\textsc{GranolaEQ}} \\
\addlinespace[2pt]
LFM & 5.35 [3.94, 7.27] & 5.78 [4.08, 8.19] & 4.18 [3.02, 5.77] & 4.14 [2.99, 5.74] \\
Ministral & 2.57 [1.95, 3.38] & 2.91 [2.02, 4.19] & 4.28 [3.33, 5.49] & 4.12 [3.18, 5.32] \\
Qwen & 3.65 [2.77, 4.81] & 6.65 [4.66, 9.49] & 11.26 [7.80, 16.26] & 11.44 [7.86, 16.65] \\
Gemma & 3.81 [2.55, 5.68] & 3.52 [2.07, 5.96] & 4.91 [3.23, 7.46] & 4.59 [2.91, 7.25] \\
\midrule
\multicolumn{5}{@{}l}{\textsc{SimpleQA}} \\
\addlinespace[2pt]
LFM & 3.27 [2.54, 4.20] & 4.13 [3.11, 5.49] & 1.31 [1.05, 1.62] & 1.25 [1.001, 1.56] \\
Ministral & 4.48 [3.64, 5.50] & 3.32 [2.63, 4.19] & 2.72 [2.22, 3.35] & 2.82 [2.27, 3.51] \\
Qwen & 4.08 [3.20, 5.22] & 3.03 [2.26, 4.05] & 4.72 [3.59, 6.21] & 4.76 [3.56, 6.36] \\
Gemma & 3.29 [2.57, 4.22] & 2.25 [1.67, 3.01] & 2.94 [2.17, 3.99] & 2.53 [1.85, 3.45] \\
\midrule
\multicolumn{5}{@{}l}{\textsc{TruthfulQA}} \\
\addlinespace[2pt]
LFM & 2.81 [1.65, 4.77] & 2.82 [1.48, 5.38] & 1.76 [1.01, 3.05] & 1.67 [0.92, 3.03] \\
Ministral & 3.06 [1.95, 4.81] & 2.70 [1.58, 4.63] & 1.79 [1.13, 2.83] & 1.48 [0.88, 2.47] \\
Qwen & 7.15 [3.81, 13.44] & 3.01 [1.38, 6.58] & 9.25 [4.46, 19.18] & 3.46 [1.49, 8.05] \\
Gemma & 2.94 [1.73, 5.00] & 1.09 [0.54, 2.19] & 2.62 [1.44, 4.77] & 1.21 [0.53, 2.76] \\
\bottomrule
\end{tabular}
\caption{Coarsening odds ratios for low versus high confidence quartiles,
with 95\% cluster-robust confidence intervals. Both specifications adjust
for relation; base controls additionally include base-answer
\textsc{GranuScore} and base-answer word count.}
\label{tab:coarsening-starting-controls}
\end{table*}

\paragraph{Sensitivity to confidence specification.}
\label{app:sensitivity-confidence}
We repeat the primary analysis and compare the lowest and highest 20\%, 25\%, 30\%, and
35\% of confidence values and fit a continuous specification
using all eligible pairs with within-group variation.
All fits adjust for relation, base-answer \textsc{GranuScore},
and base-answer word count.

For continuous confidence, we subtract each
model--dataset--relation group's mean and divide by its
standard deviation, then reverse the sign so that higher
values indicate lower confidence.
The resulting odds ratio describes a one-standard-deviation
decrease in confidence within these groups.

\autoref{tab:confidence-sensitivity} reports the results.
All controlled point estimates remain above one.
Tail-comparison odds ratios generally decrease as wider
tails include less extreme answers, with some confidence
intervals including one.
For continuous confidence, 95\% intervals exclude one in
11 of 12 model--dataset combinations under each measure;
Gemma on \textsc{TruthfulQA} is the exception.
The association therefore persists across confidence
specifications, although its strength and precision vary.

\begin{table*}[ht]
\small
        \setlength{\abovecaptionskip}{4pt}
\centering
\begin{tabular}{@{}lrrrrr@{}}
\toprule
Model & 20\% tails & 25\% tails & 30\% tails & 35\% tails & Continuous \\
\midrule
\multicolumn{6}{@{}l}{\textsc{GranolaEQ}} \\
\addlinespace[3pt]
\multicolumn{6}{@{}l}{\quad\textit{Combined metric}} \\
\quad\quad LFM & 6.58 [4.42, 9.78] & 5.78 [4.08, 8.19] & 5.23 [3.80, 7.20] & 4.64 [3.45, 6.25] & 2.28 [1.97, 2.65] \\
\quad\quad Ministral & 3.75 [2.45, 5.74] & 2.91 [2.02, 4.19] & 2.28 [1.64, 3.15] & 1.95 [1.44, 2.62] & 1.54 [1.34, 1.76] \\
\quad\quad Qwen & 8.83 [5.90, 13.21] & 6.65 [4.66, 9.49] & 5.40 [3.95, 7.39] & 4.81 [3.61, 6.41] & 2.16 [1.89, 2.47] \\
\quad\quad Gemma & 4.36 [2.37, 8.02] & 3.52 [2.07, 5.96] & 3.07 [1.94, 4.85] & 2.30 [1.54, 3.45] & 1.29 [1.10, 1.51] \\
\addlinespace[3pt]
\multicolumn{6}{@{}l}{\quad\textit{LLM judge}} \\
\quad\quad LFM & 5.00 [3.40, 7.34] & 4.14 [2.99, 5.74] & 4.03 [2.99, 5.43] & 3.77 [2.85, 4.98] & 1.91 [1.62, 2.25] \\
\quad\quad Ministral & 4.04 [3.04, 5.38] & 4.12 [3.18, 5.32] & 3.67 [2.91, 4.62] & 3.25 [2.62, 4.02] & 1.74 [1.57, 1.93] \\
\quad\quad Qwen & 13.72 [8.84, 21.28] & 11.44 [7.86, 16.65] & 8.92 [6.44, 12.36] & 7.40 [5.52, 9.92] & 2.77 [2.41, 3.20] \\
\quad\quad Gemma & 5.75 [3.35, 9.87] & 4.59 [2.91, 7.25] & 3.36 [2.27, 4.99] & 3.22 [2.24, 4.63] & 1.58 [1.37, 1.82] \\
\midrule
\multicolumn{6}{@{}l}{\textsc{SimpleQA}} \\
\addlinespace[3pt]
\multicolumn{6}{@{}l}{\quad\textit{Combined metric}} \\
\quad\quad LFM & 5.49 [3.97, 7.58] & 4.13 [3.11, 5.49] & 3.50 [2.71, 4.52] & 3.07 [2.43, 3.88] & 1.73 [1.57, 1.90] \\
\quad\quad Ministral & 3.91 [3.00, 5.10] & 3.32 [2.63, 4.19] & 2.91 [2.36, 3.59] & 2.74 [2.26, 3.32] & 1.61 [1.48, 1.76] \\
\quad\quad Qwen & 3.22 [2.32, 4.47] & 3.03 [2.26, 4.05] & 2.68 [2.06, 3.48] & 2.28 [1.80, 2.90] & 1.50 [1.36, 1.66] \\
\quad\quad Gemma & 2.20 [1.57, 3.10] & 2.25 [1.67, 3.01] & 2.41 [1.85, 3.13] & 2.26 [1.78, 2.88] & 1.31 [1.20, 1.43] \\
\addlinespace[3pt]
\multicolumn{6}{@{}l}{\quad\textit{LLM judge}} \\
\quad\quad LFM & 1.42 [1.11, 1.83] & 1.25 [1.001, 1.56] & 1.22 [1.000, 1.50] & 1.18 [0.98, 1.42] & 1.12 [1.04, 1.21] \\
\quad\quad Ministral & 3.25 [2.53, 4.17] & 2.82 [2.27, 3.51] & 2.40 [1.98, 2.92] & 2.09 [1.75, 2.50] & 1.47 [1.36, 1.58] \\
\quad\quad Qwen & 5.04 [3.62, 7.02] & 4.76 [3.56, 6.36] & 4.07 [3.15, 5.25] & 3.53 [2.80, 4.45] & 1.64 [1.49, 1.81] \\
\quad\quad Gemma & 2.83 [1.98, 4.04] & 2.53 [1.85, 3.45] & 2.33 [1.76, 3.09] & 2.11 [1.63, 2.72] & 1.32 [1.22, 1.44] \\
\midrule
\multicolumn{6}{@{}l}{\textsc{TruthfulQA}} \\
\addlinespace[3pt]
\multicolumn{6}{@{}l}{\quad\textit{Combined metric}} \\
\quad\quad LFM & 2.88 [1.44, 5.79] & 2.82 [1.48, 5.38] & 2.60 [1.47, 4.63] & 2.52 [1.51, 4.21] & 1.44 [1.17, 1.77] \\
\quad\quad Ministral & 3.03 [1.65, 5.58] & 2.70 [1.58, 4.63] & 2.20 [1.36, 3.57] & 1.78 [1.17, 2.71] & 1.37 [1.15, 1.64] \\
\quad\quad Qwen & 4.35 [1.73, 10.93] & 3.01 [1.38, 6.58] & 1.54 [0.79, 3.01] & 1.39 [0.76, 2.54] & 1.51 [1.18, 1.92] \\
\quad\quad Gemma & 1.16 [0.50, 2.70] & 1.09 [0.54, 2.19] & 1.38 [0.74, 2.60] & 1.51 [0.86, 2.66] & 1.21 [0.99, 1.49] \\
\addlinespace[3pt]
\multicolumn{6}{@{}l}{\quad\textit{LLM judge}} \\
\quad\quad LFM & 2.18 [1.07, 4.42] & 1.67 [0.92, 3.03] & 1.56 [0.91, 2.66] & 1.91 [1.17, 3.10] & 1.23 [1.02, 1.48] \\
\quad\quad Ministral & 1.92 [1.08, 3.42] & 1.48 [0.88, 2.47] & 1.25 [0.79, 1.99] & 1.16 [0.77, 1.74] & 1.20 [1.01, 1.43] \\
\quad\quad Qwen & 5.07 [1.80, 14.26] & 3.46 [1.49, 8.05] & 2.24 [1.08, 4.62] & 1.92 [0.995, 3.71] & 1.50 [1.18, 1.90] \\
\quad\quad Gemma & 1.42 [0.56, 3.60] & 1.21 [0.53, 2.76] & 1.34 [0.67, 2.70] & 1.39 [0.76, 2.55] & 1.04 [0.81, 1.32] \\
\bottomrule
\end{tabular}
\caption{Sensitivity of relation-adjusted coarsening odds ratios to confidence specification.
Brackets show 95\% cluster-robust confidence intervals. Tail specifications compare the lowest and highest
confidence fractions within each model--dataset--relation group. Continuous estimates use all eligible pairs
and describe a one-standard-deviation decrease in confidence within these groups.
All fits additionally adjust for base-answer \textsc{GranuScore} and word count.}
\label{tab:confidence-sensitivity}
\end{table*}

\end{document}